\documentclass[10pt,twocolumn,letterpaper]{article}

\usepackage[pagenumbers]{cvpr}   %

\usepackage[dvipsnames]{xcolor}

\definecolor{cvprblue}{rgb}{0.21,0.49,0.74}
\usepackage[pagebackref,breaklinks,colorlinks,allcolors=cvprblue]{hyperref}
\usepackage[T1]{fontenc}    %
\usepackage{makecell} 

\usepackage{array}
\usepackage{multirow}
\usepackage{booktabs}       %
\usepackage{float}
\usepackage{graphicx}
\usepackage{marvosym}

\usepackage{xspace}

\definecolor{myyellow}{HTML}{D3AD1D}
\definecolor{myblue}{HTML}{1E5DBC}
\definecolor{mygreen}{HTML}{5C7F11}

\usepackage{pifont}
\newcommand{\cmark}{\text{\ding{51}}}
\newcommand{\xmark}{\text{\ding{55}}}

\usepackage[capitalize]{cleveref}
\crefname{section}{Sec.}{Secs.}
\Crefname{section}{Section}{Sections}
\Crefname{table}{Table}{Tables}
\crefname{table}{Tab.}{Tabs.}

\def\paperID{1741} %
\def\confName{CVPR}
\def\confYear{2025}

\title{FreeCloth: Free-form Generation Enhances Challenging Clothed Human Modeling}

\author{
Hang Ye\textsuperscript{1} \quad Xiaoxuan Ma\textsuperscript{1, \Letter} \quad Hai Ci\textsuperscript{1} \quad Wentao Zhu\textsuperscript{1} \quad Yizhou Wang\textsuperscript{1, 2, 3, 4, \Letter} \\ 
\textsuperscript{1~}Center on Frontiers of Computing Studies,
School of Computer Science, Peking University \\
\textsuperscript{2~}Inst. for Artificial Intelligence, Peking University \quad
\textsuperscript{3~}Nat'l Eng. Research Center of Visual Technology \\ 
\textsuperscript{4~}State Key Laboratory of General Artificial Intelligence, Peking University \\ [0.12ex]
{\tt\small \{yehang, maxiaoxuan, cihai, wtzhu, yizhou.wang\}@pku.edu.cn} \quad {\tt\small $^{\textrm{\Letter}}\!$ Corresponding authors}}

\begin{document}
\twocolumn[{%
    \renewcommand\twocolumn[1][]{#1}%
    \setlength{\tabcolsep}{0.0mm} %
    \newcommand{\sz}{0.125}  %
    \maketitle
    \vspace{-1.2em}
    \begin{center}
        \newcommand{\teaserwidth}{\textwidth}
        \includegraphics[width=\linewidth]{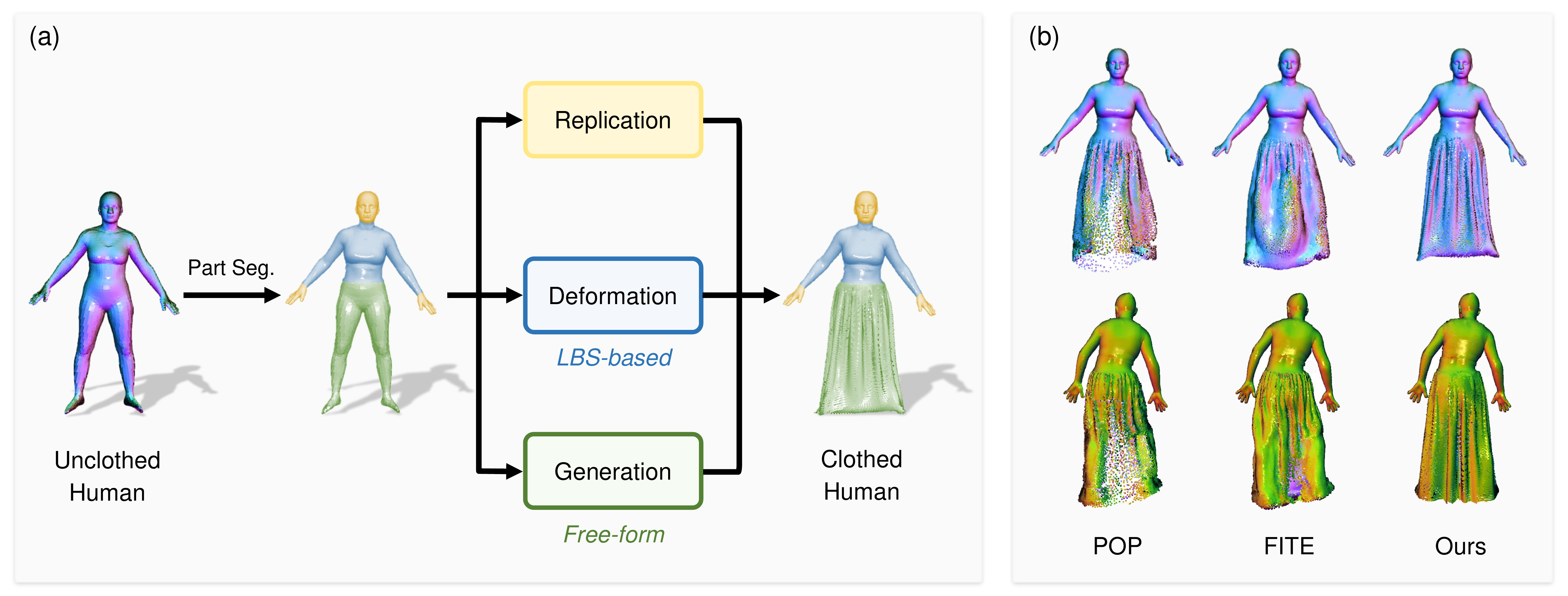}
    \vspace{-4ex}
    \captionof{figure}{
    \textbf{(a) An overview of our framework for modeling clothed humans.} Based on the specific modeling needs of different regions, we employ a dedicated strategy to handle various clothing areas. Specifically, for loose regions (\textcolor{mygreen}{green}) that are less affected by body movements and require more freedom, we propose free-form generation to enhance flexibility. For near-body clothing areas (\textcolor{myblue}{blue}), we apply LBS-based deformation, while unclothed regions (\textcolor{myyellow}{yellow}) that do not require deformation can be directly replicated. \textbf{(b) Visual comparison between prior arts (POP~\cite{ma2021pop}, FITE~\cite{lin2022fite}) and our method on challenging clothing.} Our method captures more high-fidelity details and achieves superior visual quality and realism. Code is available at \url{https://alvinyh.github.io/FreeCloth}.} 
    \label{fig:teaser}
    \end{center}%
}]

\begin{abstract}
Achieving realistic animated human avatars requires accurate modeling of pose-dependent clothing deformations. Existing learning-based methods heavily rely on the Linear Blend Skinning (LBS) of minimally-clothed human models like SMPL to model deformation. However, they struggle to handle loose clothing, such as long dresses, where the canonicalization process becomes ill-defined when the clothing is far from the body, leading to disjointed and fragmented results. 
To overcome this limitation, we propose FreeCloth, a novel hybrid framework to model challenging clothed humans. Our core idea is to use dedicated strategies to model different regions, depending on whether they are close to or distant from the body. 
Specifically, we segment the human body into three categories: unclothed, deformed, and generated. We simply replicate unclothed regions that require no deformation. For deformed regions close to the body, we leverage LBS to handle the deformation. As for the generated regions, which correspond to loose clothing areas, we introduce a novel free-form, part-aware generator to model them, as they are less affected by movements.
This free-form generation paradigm brings enhanced flexibility and expressiveness to our hybrid framework, enabling it to capture the intricate geometric details of challenging loose clothing, such as skirts and dresses.
Experimental results on the benchmark dataset featuring loose clothing demonstrate that FreeCloth achieves state-of-the-art performance with superior visual fidelity and realism, particularly in the most challenging cases. 
\end{abstract}
\vspace{-0.3cm} 
    
\section{Introduction}
\label{sec:intro}

The emergence of clothed 3D human characters, often referred to as ``digital avatars'', has swiftly evolved into a fundamental aspect across diverse industries such as gaming~\cite{kavan2011physics}, animation~\cite{he2021arch++}, virtual try-on~\cite{naik2024dress}, \etc. However, it remains an open problem to create avatars with naturally deforming clothing driven by diverse body poses, since it is difficult to capture the intricate geometry of clothing, such as wrinkles. Although conventional solutions such as rigging and skidding~\cite{baran2007automatic, feng2015avatar, liu2019neuroskinning} achieve promising results, they are highly dependent on artistic efforts and expert knowledge. To automate this process, recent studies~\cite{ma2021scale,ma2021pop,ma2022skirt, zhang2023closet} adopt a data-driven approach to learn the pose-dependent clothing deformation. Specifically, they predict local transformation in the canonical space, which is further added on top of the human body template and then driven by LBS transformation.

Nevertheless, this posing procedure often fails in terms of garments that differ greatly from the body shape and topology, especially loose clothing such as skirts and long dresses. The deformed shape is usually constrained to the minimally-clothed body, leading to split-like artifacts in modeling the long dress (see results of POP~\cite{ma2021pop} in \cref{fig:teaser}). 
This is mainly due to the poorly defined canonicalization process~\cite{ma2022skirt} in the region far away from the body, such as the area between the legs.
To alleviate this issue, recent works~\cite{ma2022skirt, lin2022fite, zhang2023closet} propose a coarse-to-fine approach for predicting deformations based on learned clothing templates. However, these approaches still confine the deformation within the LBS-based transformation, without addressing the fundamental challenge of accurately modeling complex clothing far from the body. As a result, these methods still struggle to model loose and challenging clothing accurately (see \cref{fig:teaser} (b) and \cref{fig:sota} for comparison).

In this work, we revisit the task of clothed human modeling from a novel perspective. Our pivotal insight is that integrating structural priors, \ie LBS, significantly facilitates the task, whereas relying entirely on LBS hampers flexibility. To that end, we propose FreeCloth, a hybrid framework, leveraging the complementary advantages of LBS-based and LBS-free techniques. We first conduct part segmentation to categorize surface points on the human body into three types: unclothed (\textcolor{myyellow}{yellow}), deformed (\textcolor{myblue}{blue}), and generated (\textcolor{mygreen}{green}), as shown in \cref{fig:teaser} (a).
The yellow areas represent unclothed parts (\textit{e.g.} head, hands, and feet), usually not covered by garments, which need no deformation. The blue areas represent parts close to the body, where we perform LBS-based deformation \cite{ma2021pop, ma2022skirt, lin2022fite, zhang2023closet}. The green areas indicate loose clothing regions that deviate significantly from the body and are therefore less affected by body movements. For these regions, we introduce a free-form generator to model the dynamics. Finally, we obtain a completely clothed human by merging the three branches.

\begin{table}[t]
    \centering
    \resizebox{0.75\linewidth}{!}{
    \begin{tabular}{c >{\centering\arraybackslash}p{0.6cm} >{\centering\arraybackslash}p{0.6cm} >{\centering\arraybackslash}p{0.6cm} >{\centering\arraybackslash}p{0.6cm} >{\centering\arraybackslash}p{0.6cm}}
        Method & \rotatebox{35}{\textit{w/o 2D rendering}} & \rotatebox{35}{\textit{w/o clothing template}} & \rotatebox{35}{\textit{w/o LBS field}} & \rotatebox{35}{\textit{open surface modeling}} & \rotatebox{35}{\textit{loose clothing}} \\
        \hline
        POP~\cite{ma2021pop} & \textcolor{red}{\xmark} & \textcolor{green}{\cmark} & \textcolor{green}{\cmark} & \textcolor{green}{\cmark} & \textcolor{red}{\xmark} \\
        SkiRT~\cite{ma2022skirt} & \textcolor{red}{\xmark} & \textcolor{red}{\xmark} & \textcolor{red}{\xmark} & \textcolor{green}{\cmark} & \textcolor{green}{\cmark} \\
        FITE~\cite{lin2022fite} & \textcolor{red}{\xmark} & \textcolor{red}{\xmark} & \textcolor{red}{\xmark} & \textcolor{red}{\xmark} & \textcolor{green}{\cmark} \\
        CloSET~\cite{zhang2023closet} & \textcolor{green}{\cmark} & \textcolor{red}{\xmark} & \textcolor{green}{\cmark} & \textcolor{green}{\cmark} & \textcolor{green}{\cmark} \\
        \hline
        \textbf{Ours} & \textcolor{green}{\cmark} & \textcolor{green}{\cmark} & \textcolor{green}{\cmark} & \textcolor{green}{\cmark} & \textcolor{green}{\cmark \hspace{-0.4em} \cmark} \\
        \hline
    \end{tabular}
    }
    \vspace{-1ex}
    \caption{\textbf{Comparison of our method with existing works.}}
    \label{tab:methods}
\end{table}

To guide the free-form generation of loose clothing for a posed human point cloud, we introduce structure-aware pose encoding. We extract part-based pose features from the unclothed point cloud and transform them into a pose code. The generator then predicts loose areas conditioned on this pose code and the garment type, without relying on LBS-based transformations. By prioritizing part-aware pose details over a direct global pose code, this approach ensures a closer alignment between the generated clothing and the given poses, thereby enhancing the high fidelity and realism of the results. As demonstrated in \cref{fig:teaser} (b), thanks to the flexibility of the generator, our hybrid framework successfully generates realistic and intricate wrinkles for loose dresses and skirts, eliminating pant-like artifacts.

We conduct evaluations on long dresses and skirts with diverse lengths, styles, and tightness levels. Experimental results on the benchmark dataset featuring loose clothing demonstrate that our method achieves state-of-the-art (SOTA) performance with superior visual fidelity and realism, particularly in the most challenging cases. To the best of our knowledge, we are the first to leverage free-form generation to tackle learning-based clothed human modeling. Being single-staged and end-to-end, our simple yet effective paradigm significantly enhances the expressiveness of clothed avatars. It excels at capturing fine details of loose clothing without the need to render 2D positional maps~\cite{ma2021pop, ma2022skirt, lin2022fite}, extract subject-specific clothing templates~\cite{ma2022skirt, lin2022fite}, or learn continuous LBS fields~\cite{ma2022skirt, lin2022fite}, while also offering the flexibility to model open surfaces. We emphasize the key strengths of our free-form paradigm when compared to recent SOTA methods in \cref{tab:methods}. 

Our main contributions are summarized as follows: 
\begin{itemize}
\item We propose a novel perspective on hybrid modeling for clothed humans, allowing for customized modeling of different body areas based on part segmentation.
\item We propose a free-form generator with structure-aware pose encoding to model loose clothing that enhances flexibility and expressiveness.
\item Our hybrid framework, FreeCloth, merging the merits of LBS deformation and free-form generator, delivers SOTA performance and enhanced visual fidelity and realism, especially in the most challenging cases.
\end{itemize}

\section{Related Work}
\label{sec:formatting}

\subsection{3D Representations for Clothed Human}

\noindent\textbf{Surface Meshes} are efficient and compatible representations for modeling 3D clothed humans. Prevailing approaches represent clothing either as a deviation from the body~\cite{bhatnagar2019multi, burov2021dynamic, ma2020cape, neophytou2014layered, tiwari2020sizer, su2022deepcloth} or as a separate layer~\cite{guan2012drape, lahner2018deepwrinkles, gundogdu2019garnet, patel2020tailornet}. Nonetheless, the fixed topology of meshes struggles to generalize across varying clothing types. 

\vspace{0.5em}

\noindent\textbf{Neural Implicit Field} offers more topological flexibility~\cite{mescheder2019occupancy, park2019deepsdf} and is promising for 
reconstructing or animating clothed humans~\cite{xiu2022icon, saito2019pifu, deng2020nasa, chen2021snarf, saito2021scanimate, ma2022skirt, qian2022unif}. However, extracting the surface from the field is computationally expensive, limiting its practical use. Another line of works~\cite{su2021anerf, peng2021neural, weng2022humannerf} optimize neural radiance fields (NeRF)~\cite{mildenhall2021nerf} from 2D human images but lacks explicit geometry for accurate pose control in animation.
Recently, 3D Gaussian Splatting~\cite{3dgs} is introduced to improve real-time rendering with high visual fidelity. Several studies~\cite{pang2023ash, li2023animatable, moreau2023human, kocabas2023hugs, jung2023deformable, zheng2023gps} employ this explicit modeling technique to represent textured human models. 

\vspace{0.5em}

\noindent\textbf{Hybrid Approaches} emerge recently. DMTet~\cite{shen2021dmtet, gao2020learning} consists of an explicit tetrahedral grid and an implicit distance field. TeCH~\cite{huang2023tech} and HumanNorm~\cite{huang2024humannorm} explore the potential of DMTet in generating high-fidelity clothed humans with enhanced geometric details.

\vspace{0.5em}

\noindent\textbf{Point Clouds} enjoy efficiency as well as flexibility. Nevertheless, it's still an open challenge to generate high-resolution point clouds with fine geometric details. Prior works~\cite{bednarik2020shape, deng2020better, deprelle2019learning, groueix2018papier, ma2021scale} group points into patches to model the clothing but suffers from inter-patch discontinuity. POP~\cite{ma2021pop} further improves this by introducing fine-grained features with UV maps. Another line of works~\cite{zhang2023closet, prokudin2023dpf, zakharkin2021point} focus on eliminating the ``seaming'' artifact.  We use point clouds as our representation, as they offer greater topological flexibility and faster inference speeds~\cite{ma2021pop, zakharkin2021point} than meshes and implicit fields. 

\subsection{Animating Clothed Human Avatars} 
LBS is a predominant technique in animating human avatars, enabling the rigid transformation of the surface point in correspondence with the articulated movements of the underlying skeleton~\cite{baran2007automatic, loper2023smpl, pavlakos2019smplx}. 

\vspace{0.5em}
\noindent\textbf{LBS-based Animation.} We roughly classify LBS-based methods into explicit and implicit methods. Explicit methods deal with explicit 3D data like meshes~\cite{ma2020cape} and point clouds~\cite{ma2021pop, zhang2023closet, ma2021scale}. A common approach is to predict local transformation~\cite{ma2021pop, zhang2023closet, ma2021scale} relative to pre-defined skinning weights on SMPL~\cite{loper2023smpl}. Implicit methods further extend LBS to implicit fields. Skinning weights can be either learned from data~\cite{chen2021snarf, saito2021scanimate, ma2022skirt, kant2023invertible}, obtained by nearest neighbor~\cite{bhatnagar2020loopreg} or diffusion~\cite{lin2022fite}. However, it is non-trivial to get an accurate skinning field without direct supervision or sufficient data. Due to the inherent rigid transformation, LBS-based methods are severely limited to tight clothing.

\vspace{0.5em}
\noindent\textbf{LBS-free Animation.} A relevant work DPF~\cite{prokudin2023dpf} escapes LBS by directly optimizing a smooth deformation field. DPF generates visually impressive results but requires frame-wise optimization, restricting its practicality.

In this work, we propose a hybrid approach to model clothed humans, which can better exploit the complementary advantages of LBS-based and LBS-free approaches. 

\section{Method}

\begin{figure*}[t]
  \centering
  \includegraphics[width=1\linewidth]{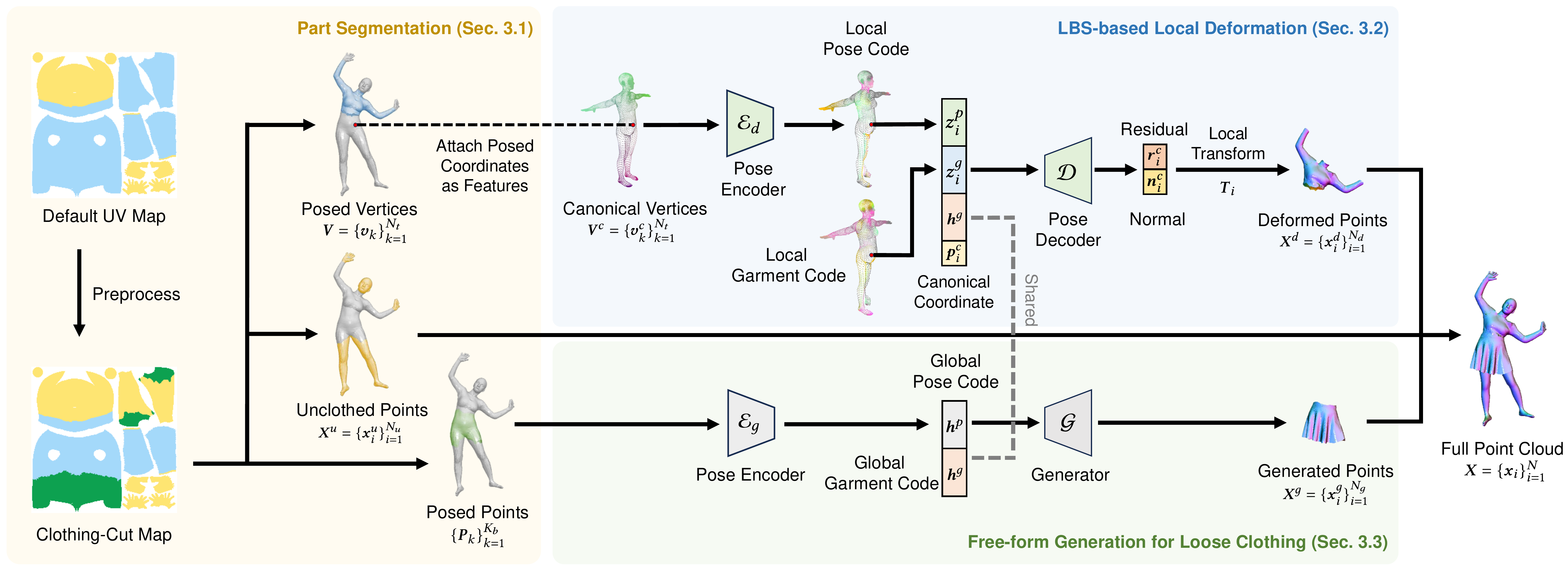}
  \vspace{-3ex}
  \caption{\textbf{Overview of our hybrid framework FreeCloth.} Given an unclothed and posed body, and a specific garment type, our goal is to create a realistic clothed human. We first segment the human parts into three different regions (\cref{sec:cut}): unclothed parts (\textcolor{myyellow}{yellow}) need no deformation, deformed parts (\textcolor{myblue}{blue}), and generated parts (\textcolor{mygreen}{green}). The hybrid framework comprises two essential modules: (1) an LBS-based local deformation network (\cref{sec:deform}) to obtain pose-dependent deformed points $\boldsymbol{X}^d$ that are close to the human body, and (2) a free-form generator that focuses on generating the more loose clothing regions $\boldsymbol{X}^g$ (\cref{sec:hybrid}). By merging the unclothed, deformed, and generated points, we ultimately obtain the complete point cloud of a clothed human $\boldsymbol{X}$. } 
  \label{fig:pipeline}
\end{figure*}

Our objective is to dress an unclothed and posed human body with a specific clothing type and create a realistic clothed human. The overall pipeline is illustrated in \cref{fig:pipeline}. Considering the varying impact of body movements on different regions of clothing, we propose a novel hybrid framework that combines three distinct strategies to model these regions, \ie unclothed, deformed, and generated. First, we identify these three types of regions to create a clothing-cut map (\cref{sec:cut}). Then we propose two essential modules to model the deformed and generated parts accordingly: (1) an LBS-based local deformation network (\cref{sec:deform}) to model near-body clothing deformation, and (2) a free-form generation module that focuses on handling the more distant clothing regions (\cref{sec:hybrid}). Finally, we describe the training strategy in \cref{sec:training}.

\subsection{Human Part Segmentation}  \label{sec:cut}

While the free-form generator liberates the constraint of LBS-based deformation and enhances the expressiveness, our hybrid design introduces an important question: \textit{how to automatically determine whether a point on the body surface should be deformed or generated?} To address this, we compute a \textbf{garment-specific clothing-cut map} to explicitly segment the human body into distinct regions, guiding our modules to handle different parts exclusively.

We first locate the exposed areas unaffected by garment coverage, such as the head, hands, and feet, which do not undergo deformations. Let $\boldsymbol{X}^u = \{\boldsymbol{x}_i^u\}_{i=1}^{N_u}$ represent these unclothed body points. Then, our key design is to segment the regions that are occluded by loose clothing, \eg, skirts or dresses, and disable LBS-based deformation within these areas. This is achieved utilizing the foundation model SAM~\cite{kirillov2023sam} to segment the loose parts from the rendered normal maps. We then back-project the detected regions into 3D space, effectively identifying the loose areas. This process also uncovers the remaining unclothed regions, such as parts of the legs not covered by a skirt, which are merged into $\boldsymbol{X}^u$, corresponding to the \textcolor{myyellow}{yellow} region in \cref{fig:pipeline}. Please refer to the supplementary material for details of defining the garment-specific clothing-cut map. Then, for near-body regions denoted in \textcolor{myblue}{blue}, we perform the LBS-based local deformation (\cref{sec:deform}), while for modeling the loose regions denoted in \textcolor{mygreen}{green}, we employ the free-form generation (\cref{sec:hybrid}).

\subsection{LBS-based Local Deformation} \label{sec:deform}
Given that clothing near the body surface is more influenced by body movements, we can leverage the body structural priors to better guide the deformation of clothing in these areas using LBS provided by a parametric human model, \ie SMPL-X~\cite{pavlakos2019smplx} used in this work. 
Given a posed and unclothed body model, we denote the posed vertices as $\boldsymbol{V}=\{\boldsymbol{v}_k\}_{k=1}^{N_t}$, where $N_t$ is the number of vertices. We define the corresponding vertices in the canonical space as $\boldsymbol{V}^c=\{\boldsymbol{v}^c_k\}_{k=1}^{N_t}$.
Unless otherwise stated, the superscript letters $c$ represent ``canonical'' in the following notation. 

\vspace{1.4em}
\noindent\textbf{Local Pose Code.} To model the pose-dependent deformation of the clothing, a naive way is to condition the deformation on a single pose encoding~\cite{saito2021scanimate,chen2021snarf}. However, later works point out that fine-grained per-point geometric pose encodings can serve as a better pose condition~\cite{ma2021pop,ma2022skirt,zhang2023closet}. Therefore, following CloSET~\cite{zhang2023closet}, we employ PointNet++~\cite{qi2017pointnet++} to extract multi-scale local pose feature $\boldsymbol{\phi}_k^p \in \mathbb{R}^{M_p}$ for each body vertex $\boldsymbol{v}_k$ (see \cref{fig:pipeline} for illustration) as defined in \cref{eq:local_pose}, where $M_p$ denotes the number of feature channels.
\begin{equation}
\label{eq:local_pose}
    \{\boldsymbol{\phi}_k^p\}_{k=1}^{N_t} = \mathcal{E}_d(\boldsymbol{V}^c, \boldsymbol{V}).
\end{equation}
Note that we treat the \textit{canonical} vertices $\boldsymbol{V}^c$ as a point cloud, where the coordinates of each \textit{posed} vertex are regarded as the features of each point. These features are then used as inputs to the PointNet++ $\mathcal{E}_d$. To obtain continuous local pose code for any point $\boldsymbol{p}_i$ located on the body surface manifold, we diffuse the local pose feature $\boldsymbol{\phi}_k^p$ on the body surface by applying barycentric interpolation. Given the barycentric coordinates $\boldsymbol{b}_i = [b_{i1}, b_{i2},b_{i3}]$ and the associated vertex indices $\boldsymbol{s}_i = [s_{i1}, s_{i2}, s_{i3}]$ of a body surface, we obtain its local pose code $\boldsymbol{z}_i^p \in \mathbb{R}^{M_p}$ as follows:
\begin{equation} \label{eq:bary_inter}
    \boldsymbol{z}_i^p = \sum\limits_{j=1}^3 (b_{ij} \cdot \boldsymbol{\phi}_{s_{ij}}^p).
\end{equation}

\vspace{0.3em}
\noindent\textbf{Garment Code.} 
To control the clothing type, we introduce a spatial-aligned local garment code $\boldsymbol{\phi}_k^g \in \mathbb{R}^{M_g}$ for each canonical vertex on the body surface following~\cite{ma2021pop}. Similar to the continuous local pose code $\boldsymbol{z}_i^p$, we apply barycentric interpolation to convert the discrete code into a continuous garment code $\boldsymbol{z}_i^g \in \mathbb{R}^{M_g}$ for any body surface point $\boldsymbol{p}_i$, as defined by \cref{eq:bary_inter_garment}. The local garment code $\boldsymbol{\phi}_k^g$ is learned in an auto-decoding~\cite{park2019deepsdf} manner and it is shared across all human poses. 
\begin{equation} \label{eq:bary_inter_garment}
    \boldsymbol{z}_i^g = \sum\limits_{j=1}^3 (b_{ij} \cdot \boldsymbol{\phi}_{s_{ij}}^g).
\end{equation}
In addition to the local garment code $\boldsymbol{z}_i^g$, we also introduce a global garment code $\boldsymbol{h}^g \in \mathbb{R}^{M_g}$, which is shared with the one used in our free-form generation module. This ensures consistency in the types of clothing generated by both modules. More details about the global garment code $\boldsymbol{h}^g$ will be discussed in the next section.

\vspace{0.5em}
\noindent\textbf{LBS-based Local Deformation.}  
For any query point $\boldsymbol{p}_i$ located on the posed body surface manifold, we concatenate its local pose code $\boldsymbol{z}_i^p$, the corresponding canonical point $\boldsymbol{p}_i^c$, and the garment codes $\boldsymbol{z}_i^g$, $\boldsymbol{h}^g$ together as a feature vector and pass through a pose decoder $\mathcal{D}$~\cite{ma2021pop} to predict deformation in the canonical space:
\begin{equation}
[\boldsymbol{r}_i^c, \boldsymbol{n}_i^c] = \mathcal{D}(\boldsymbol{z}_i^p, \boldsymbol{z}_i^g, \boldsymbol{h}^g, \boldsymbol{p}_i^c),
\end{equation}
where $\boldsymbol{r}_i^c$, $\boldsymbol{n}_i^c$ denote per-vertex displacement and normal, respectively. See \cref{fig:pipeline} for the workflow.

We then add the predicted deformation to the canonical points $\boldsymbol{p}_i^c$ and then apply a local transformation $\boldsymbol{T}_i$ using LBS weight to obtain the deformed points $\boldsymbol{x}_i^d$ in the posed space, following the common LBS-based deformation practice in recent works~\cite{ma2021pop, ma2022skirt, lin2022fite, zhang2023closet}: 
\begin{equation} \label{eq:transf}
    \boldsymbol{x}_i^d = \boldsymbol{p}_i + \boldsymbol{T}_i \cdot \boldsymbol{r}_i^c  = \boldsymbol{T}_i \cdot (\boldsymbol{p}_i^c+\boldsymbol{r}_i^c).
\end{equation}
Likewise, the predicted normal $\boldsymbol{n}_i^c$ is transformed to $\boldsymbol{n}_i^d$ accordingly via the rotation component $\boldsymbol{R}_i$ of the transformation $\boldsymbol{T}_i$:
\begin{equation}
    \boldsymbol{n}_i^d = \boldsymbol{R}_i \cdot \boldsymbol{n}_i^c,
\end{equation}
where $\boldsymbol{T}_i$ is computed using barycentric interpolation of the LBS-induced bone transformation predefined in the SMPL-X~\cite{pavlakos2019smplx} body model. Now we obtain a deformed point cloud $\boldsymbol{X}^d = \{\boldsymbol{x}_i^d\}_{i=1}^{N_d}$ with its normals $\boldsymbol{N}^d = \{\boldsymbol{n}_i^d\}_{i=1}^{N_d}$ that captures the pose-dependent clothing deformation. For clarity, we omit the normal notation in \cref{fig:pipeline}.

\subsection{Free-form Generation for Loose Clothing} \label{sec:hybrid}

Although LBS-based deformation works well for points that are close to the body surface, it encounters challenges when dealing with points that are farther away from the body. This is caused by the ill-defined canonicalization process~\cite{ma2022skirt} for those points, resulting in difficulty in estimating the nonrigid transformations. This limitation becomes particularly evident when handling loose clothing such as skirts, as observed in the results of POP~\cite{ma2021pop} in \cref{fig:sota}, where the skirts are torn apart. Given the unique characteristics of skirts, it becomes infeasible to accurately model points that are distant from the body, such as those located between the legs, solely relying on the LBS deformation. \textbf{Conversely, it should be considered as a separate and flexible part}. Note that in another line of work~\cite{guan2012drape, lahner2018deepwrinkles, gundogdu2019garnet, santesteban2019learning, patel2020tailornet}, although these approaches model the garment as a separate layer, they still rely on the LBS to manipulate the garment deformation. In contrast, we propose a free-form approach to modeling loose garments, getting rid of LBS entirely.

\vspace{1.4em}
\noindent\textbf{Structure-aware Pose Encoding.} Conceptually, generating loose garments given a specific pose can be interpreted as a task of point cloud completion. To better condition on human poses, we modify the off-the-shelf SpareNet~\cite{xie2021style} to be structure-aware as our generator. We first segment the human body surface into $K_b$ semantic parts and uniformly sample posed points $\{\boldsymbol{P}_k\}_{k=1}^{K_b}$ from these parts (with a slight abuse of notation). We also replace the original PointNet~\cite{qi2017pointnet} with PointNet++~\cite{qi2017pointnet++} as pose encoder $\mathcal{E}_g$ to extract part-wise local features $\boldsymbol{h}^p_k$ for each part, which are then fused to produce a part-based pose code $\boldsymbol{h}^p \in \mathbb{R}^{M_p}$.  Note that $\boldsymbol{h}^p$ is shared for all vertices, while the local pose code $\boldsymbol{z}_i^p$ in \cref{sec:deform} is unique for each vertex. In contrast to directly extracting global features from the overall point cloud, this structure-aware design better captures the correlation between the loose garment and the underlying skeleton. 
\begin{equation}
    \boldsymbol{h}^p = \text{Max-Pooling}(\{\mathcal{E}_g(\boldsymbol{P}_k)\}_{k=1}^{K_b}).
\label{eq:generate_enc}
\end{equation}

Given the part-based pose code $\boldsymbol{h}^p$ and the garment code $\boldsymbol{h}^g$, we generate a set of points $\boldsymbol{X^g}$ \textbf{in the posed space}: 
\begin{equation}
    \boldsymbol{X^g} = \{\boldsymbol{x}_i^g\}_{i=1}^{N_g} = \mathcal{G}(\boldsymbol{h}^p, \boldsymbol{h}^g),
\label{eq:generate}
\end{equation}
where $\boldsymbol{h}^g$ is used to control the garment type and is shared between the two modules to ensure consistency.
Note that the LBS transformation is not involved in this process, which circumvents the limitations of estimating non-rigid deformation of loose clothing, hence enabling ``free-form'' generation. For detailed architecture, please refer to \cref{supp:sec:supp_arch} in the supplementary material.

\begin{table*}[h]
\centering
\caption{\textbf{Quantitative comparison of different methods on the ReSynth~\cite{ma2021pop} dataset for each subject.} We report FID scores for the rendered multi-view normal maps, along with MSE errors (in units of $10^{-2}$) between these maps and the GT normals. The best results are highlighted in \textbf{bold}, and the second best are \underline{underlined}. The subject IDs are listed in descending order based on the looseness of the clothing. Notably, the advantages of our method become more pronounced for the most challenging cases.}
\label{tab:resynth_main}
\resizebox{\linewidth}{!}{ %
\begin{tabular}{@{}c|cc|cccccccccc@{}}
\toprule
Subject                     & \multicolumn{2}{c|}{All} & \multicolumn{2}{c}{felice-004} & \multicolumn{2}{c}{janett-025}  & \multicolumn{2}{c}{christine-027} & \multicolumn{2}{c}{anna-001}     & \multicolumn{2}{c}{beatrice-025} \\ \midrule
\multicolumn{1}{c|}{Metric}    & FID~$\downarrow$   & \multicolumn{1}{c|}{MSE~$\downarrow$} & FID~$\downarrow$   & \multicolumn{1}{c}{MSE~$\downarrow$} & FID~$\downarrow$   & \multicolumn{1}{c}{MSE~$\downarrow$} & FID~$\downarrow$    & \multicolumn{1}{c}{MSE~$\downarrow$}  & FID~$\downarrow$  & \multicolumn{1}{c}{MSE~$\downarrow$}  & FID~$\downarrow$ & MSE~$\downarrow$ \\ \midrule
\multicolumn{1}{c|}{POP~\cite{ma2021pop}}  & 57.87 & \multicolumn{1}{c|}{2.88} & 66.43 & \multicolumn{1}{c}{5.80}  & 52.55  & \multicolumn{1}{c}{2.02} & 61.09  & \multicolumn{1}{c}{2.64} & 51.48  & \multicolumn{1}{c}{2.05} & 57.82 & 1.86 \\ 
\multicolumn{1}{c|}{SkiRT~\cite{ma2022skirt}} & 53.32 & \multicolumn{1}{c|}{2.72} & 63.27 & \multicolumn{1}{c}{5.70} & 48.23  & \multicolumn{1}{c}{2.03} & 55.84  & \multicolumn{1}{c}{\underline{2.44}} & 50.26 & \multicolumn{1}{c}{\textbf{1.81}} & 54.00 & \textbf{1.60} \\ 
\multicolumn{1}{c|}{FITE~\cite{lin2022fite}} & \underline{39.02} & \multicolumn{1}{c|}{\underline{2.70}} & \textbf{38.61} & \multicolumn{1}{c}{\textbf{5.09}} & \underline{35.81} & \multicolumn{1}{c}{2.09} & \underline{40.83} & \multicolumn{1}{c}{2.52} & \textbf{38.21} & \multicolumn{1}{c}{1.97} & \underline{41.62} & 1.82 \\
\multicolumn{1}{c|}{\textbf{Ours}} & \textbf{37.75} & \multicolumn{1}{c|}{\textbf{2.61}} & \underline{42.41}  & \multicolumn{1}{c}{\underline{5.24}} & \textbf{27.95} &  \multicolumn{1}{c}{\textbf{1.92}} & \textbf{37.43} & \multicolumn{1}{c}{\textbf{2.35}} & \underline{39.63} &  \multicolumn{1}{c}{\underline{1.89}} & \textbf{41.24} & \underline{1.68} \\ 
\bottomrule
\end{tabular}}
\end{table*}

\subsection{Training} \label{sec:training}
Our method is trained in an end-to-end manner, where the networks and the global garment codes are jointly optimized using the loss function defined below:
\begin{equation}
\small
\resizebox{\linewidth}{!}{$
\mathcal{L} = \lambda_{cd} \mathcal{L}_{cd} + \lambda_n \mathcal{L}_n + \lambda_{rd} \mathcal{L}_{rd} + \lambda_{rg} \mathcal{L}_{rg} + \lambda_{col} \mathcal{L}_{col}.
$}
\end{equation}

\vspace{1.4em}
\noindent\textbf{Reconstruction Losses.} Following previous works~\cite{ma2021pop, ma2022skirt, zhang2023closet}, we employ the normalized Chamfer distance $\mathcal{L}_{cd}$ to minimize the bi-directional distances between the predicted full point cloud $\boldsymbol{X}$ and the ground-truth (GT) human point cloud:
\begin{equation}
\small
    \mathcal{L}_{cd} = \dfrac{1}{N}\sum\limits_{i=1}^N \min\limits_j \|\boldsymbol{x}_i - \hat{\boldsymbol{x}}_j\|_2^2 + \dfrac{1}{M}\sum\limits_{j=1}^M \min\limits_i \|\boldsymbol{x}_i - \hat{\boldsymbol{x}}_j\|_2^2,
\label{eq:cd}
\end{equation}
where $\hat{\boldsymbol{x}}_j$ is the point sampled from the surface of the GT scan, $N$ and $M$ denote the number of the predicted and GT points, respectively. And the normal loss $\mathcal{L}_{n}$ is calculated as the average $\mathcal{L}_{1}$ distance between the predicted normal and its nearest counterpart in the GT point cloud:
\begin{equation}
    \mathcal{L}_n = \dfrac{1}{N}\sum\limits_{i=1}^N \|\boldsymbol{n}_i - \hat{\boldsymbol{n}}(\mathop{\arg\min}\limits_{\hat{\boldsymbol{x}}_j} d(\boldsymbol{x}_i,\hat{\boldsymbol{x}}_j))\|_1,
\label{eq:nml}
\end{equation}
where $\hat{\boldsymbol{n}}(\cdot)$ represents the normal of a GT point cloud and $\boldsymbol{n}_i$ denotes the estimated normal generated by our model.

\vspace{1.4em}
\noindent\textbf{Regularization Losses.} To constraint the deformed points not far away from the body, we introduce a regularization term $\mathcal{L}_{rd}$ to penalize the $\mathcal{L}_2$-norm of the pose-dependent displacement $\boldsymbol{r}_i^c$. In addition, the local and the global garment codes $\{\boldsymbol{z}^g, \boldsymbol{h}^g\}$ are regularized by their $\mathcal{L}_2$-norm:
\begin{equation}
\small
    \mathcal{L}_{rd} = \dfrac{1}{N_d} \sum\limits_{i=1}^{N_d} \|\boldsymbol{r}_i^c\|_2^2, \quad \mathcal{L}_{rg} = \sum\limits_{i=1}^{N_d} \dfrac{1}{N_d} \|\boldsymbol{z}_i^g\|_2^2+ \|\boldsymbol{h}^g\|_2^2.
\end{equation}

\vspace{1.4em}
\noindent\textbf{Collision Loss.} Drawing inspiration from the literature on garment animation~\cite{santesteban2021self, shao2023towards, lee2023multi}, we propose a collision loss to prevent intersections between the clothing and the underlying body, which is computed using the following formula:
\begin{equation}
\small
    \mathcal{L}_{c} = \dfrac{1}{N_g} \sum\limits_{j=1}^{N_g} \max\{\epsilon - d(\boldsymbol{x}^g_j), 0\},
\end{equation}
where $d(\boldsymbol{x}^g_j)$ represents the signed distance function (SDF) value of the generated points relative to the underlying body field, and $\epsilon$ is a predefined threshold that regulates the minimum distance between the body and the garment. 

\section{Experiments}

\noindent\textbf{Datasets.}  We train and evaluate our method and baselines on the ReSynth~\cite{ma2021pop} dataset, which is a synthetic dataset capturing clothed human subjects with intricate geometric details and complex pose-dependent clothing deformation. We use the official training and test split as~\cite{ma2021pop}. Similar to SkiRT~\cite{ma2022skirt}, our main focus in this study lies in accurately modeling loose clothing. Our evaluation centers on five subjects adorned in skirts and dresses of various styles, lengths, and tightness levels. 

\vspace{1.4em}
\noindent\textbf{Baseline.} To evaluate the representation power of our model, we compare it with the SOTA point-based methods (\cref{sec:compare_sota}): POP~\cite{ma2021pop}, SkiRT~\cite{ma2022skirt} and FITE~\cite{lin2022fite}. Note that as CloSET~\cite{zhang2023closet} is not open-source, we are unable to compare it with our method.

\vspace{1.4em}
\noindent\textbf{Metrics.} 
As noted in prior work~\cite{prokudin2023dpf, lin2022fite}, conventional regression-based metrics like Chamfer Distance do not accurately reflect model performance. Instead, we follow approaches in 3D human reconstruction~\cite{xiu2022icon, xiu2022econ} by computing Mean Squared Error (MSE) between rendered multi-view normal maps from the point cloud and the GT. Additionally, since our avatar modeling is generative, we employ Fréchet Inception Distance (FID~\cite{heusel2017fid}) for evaluation following Chupa~\cite{kim2023chupa}. To further assess visual quality, we conduct a perceptual study with 50 volunteers. We also utilize the GPT-4o model~\cite{achiam2023gpt4} to select the best results from all methods. Details of the perceptual study and discussions of the evaluation metrics are provided in \cref{supp:sec:supp_perceptual} and \cref{supp:sec:supp_metric} of  the supplementary materials.

\begin{figure}[t]
  \centering
  \includegraphics[width=1\linewidth]{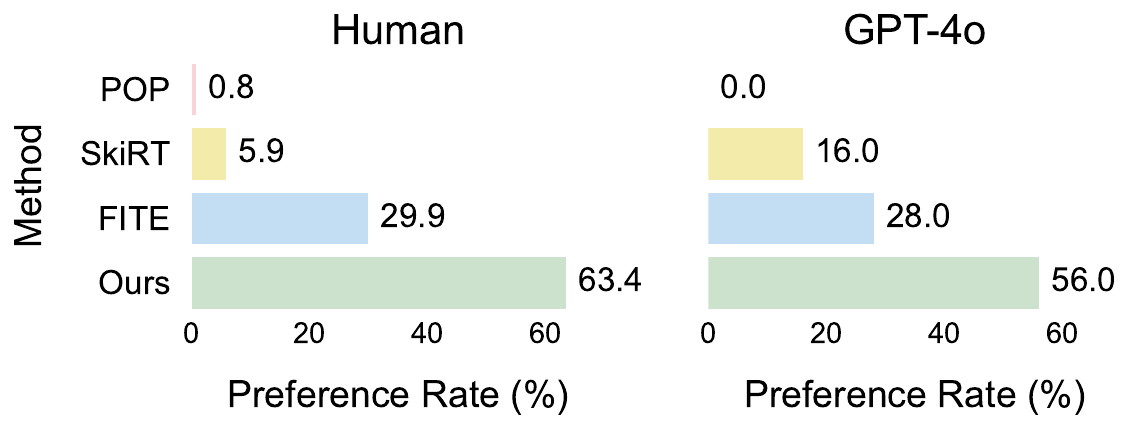}
  \vspace{-4.5ex}
  \caption{\textbf{Perceptual study results.} Across all examples, $63.4\%$ of human users prefer our method over the baselines. Additionally, our model receives $56.0\%$ of the votes from the GPT-4o model~\cite{achiam2023gpt4}. These results highlight the significant superiority of our approach, particularly in handling the most challenging clothing.}
  \label{fig:user}
\end{figure}

\subsection{Comparison with the State-of-the-arts} \label{sec:compare_sota}

\begin{figure*}[t]
  \centering
  \includegraphics[width=1\linewidth]{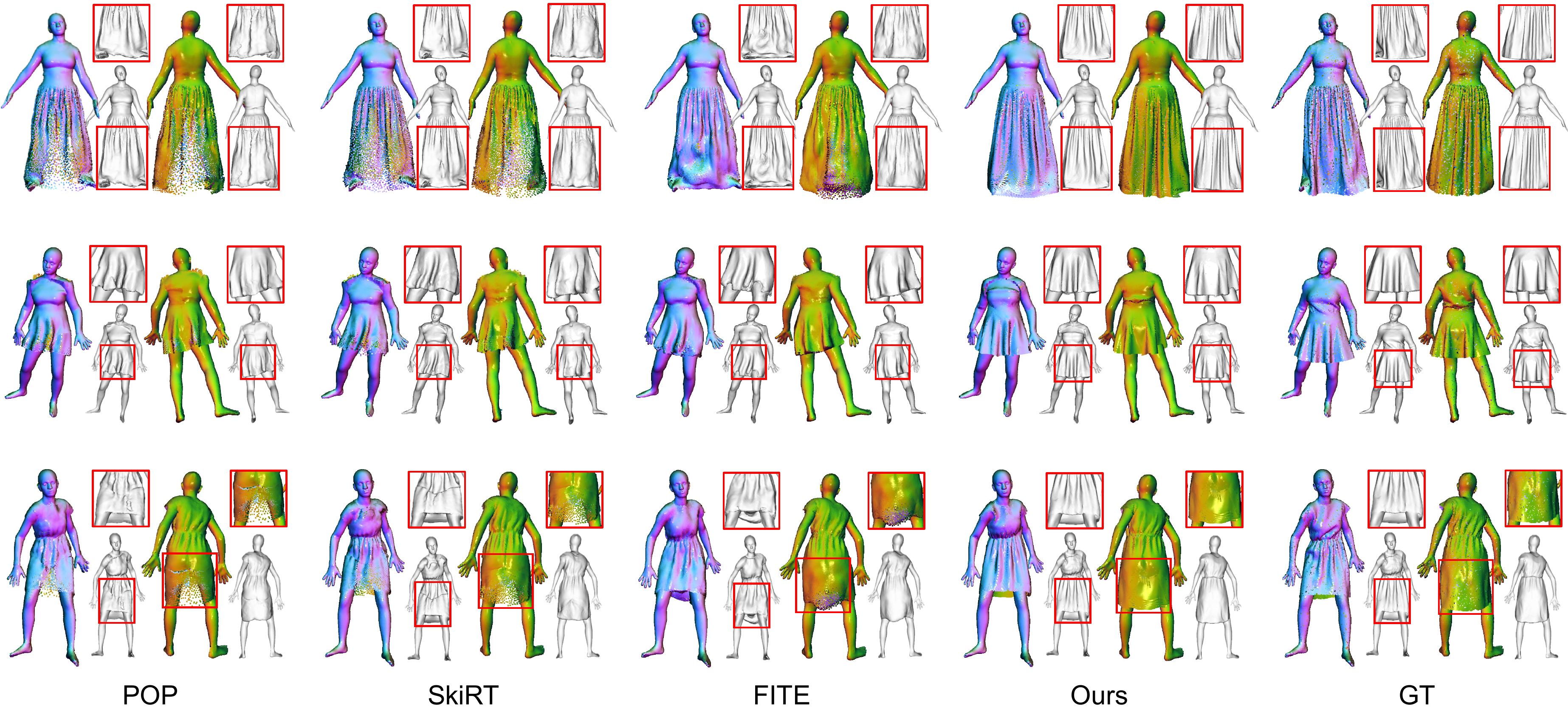}
  \vspace{-3ex}
  \caption{\textbf{Qualitative comparison between baselines and our method for modeling loose clothing.} Subject IDs from top to bottom: ``felice-004'', ``janett-025'' and ``christine-027''. Best viewed zoomed-in on a color screen.}
  \label{fig:sota}
  \vspace{-2.5ex}
\end{figure*}

\noindent\textbf{Quantitative Evaluation.} \cref{tab:resynth_main} presents the FID scores and measured MSE errors for the rendered multi-view normal maps. These metrics effectively characterize visual quality while maintaining proximity to the reference image. Our method achieves SOTA performance, surpassing other baselines with the lowest FID scores and MSE errors. This demonstrates that hybrid modeling enhances performance, particularly for loose skirts (\eg, subject janett-025). 

\cref{fig:user} shows our perceptual study, where $63.4\%$ of participants prefer the results produced by our method due to superior visual quality and closer resemblance to the GT. In comparison, FITE~\cite{lin2022fite}, SkiRT~\cite{ma2022skirt}, and POP~\cite{ma2021pop} receive $29.9\%$, $5.9\%$, and $0.8\%$ of the votes, respectively.  These findings are further supported by GPT-4o~\cite{achiam2023gpt4},  which shows $56\%$ preference for our method, aligning with the human study. Our advantage is particularly pronounced for challenging cases with loose clothing, where over $85\%$ of human evaluators favor our method for the two most difficult skirts (a detailed breakdown of the votes is available in \cref{tab:user_study_subject}).  This highlights the effectiveness of our hybrid approach in modeling loose clothing. For tighter skirts, our model performs on par with FITE, which also generates satisfactory results. However, FITE exhibits an ``open-surface'' artifact, which is not visible in the study’s front-facing renderings. We will discuss this limitation in detail later. Refer to the supplementary materials for visualization results.

\vspace{1em}
\noindent\textbf{Qualitative Results.} 
We present the qualitative results with zoomed-in details in \cref{fig:sota}. To perform a holistic evaluation of the 3D geometry, we employ Poisson reconstruction~\cite{kazhdan2013poisson} to convert the point-based representation into a triangular mesh. As can be seen, due to the inherent flaw of LBS posing, POP~\cite{ma2021pop} suffers from the ``split'' artifacts for skirts and dresses. In addition, the distribution of points is severely non-uniform, lacking realistic details such as wrinkles. SkiRT~\cite{ma2022skirt} mitigates this issue to some extent, but the results remain unsatisfactory. FITE~\cite{lin2022fite} achieves a more uniform point density after resampling on the clothing template. However, it introduces unnatural, overly bent wrinkles in long dresses due to the poorly defined LBS process. Moreover, the coarse-to-fine refinement fails to capture intricate details, often leading to noisy surfaces and loss of sharp structures (see meshing results in \cref{fig:sota}). Our method significantly outperforms the baselines in visual quality. Leveraging an LBS-free generation module, our approach effectively handles the complex, loose regions of skirts and dresses. This results in natural, high-fidelity details that closely resemble the GT, along with smooth and densely distributed points, demonstrating the representative power of our hybrid framework. 

\begin{figure}[t]
  \centering
  \includegraphics[width=0.95\linewidth]{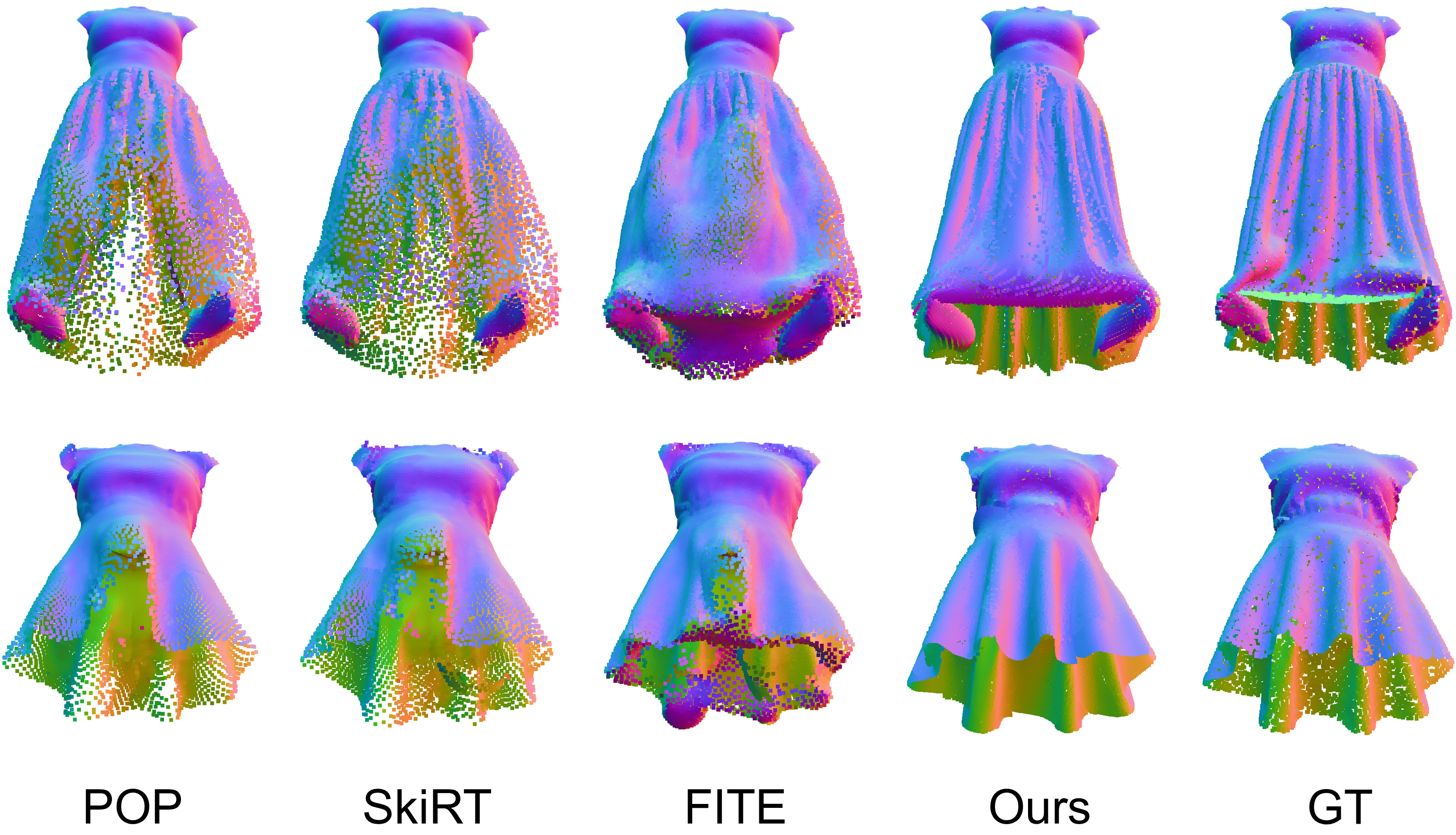}
  \caption{\textbf{Visualization results of loose clothing.} Our model effectively avoids redundant points on the open surface of loose skirts, a limitation in FITE~\cite{lin2022fite}, and generates more accurate geometry than POP~\cite{ma2021pop} and SkiRT~\cite{ma2022skirt}.}
  \label{fig:open_surface}
  \vspace{-2.5ex}
\end{figure}

\begin{figure*}[t]
  \centering
  \includegraphics[width=0.95\linewidth]{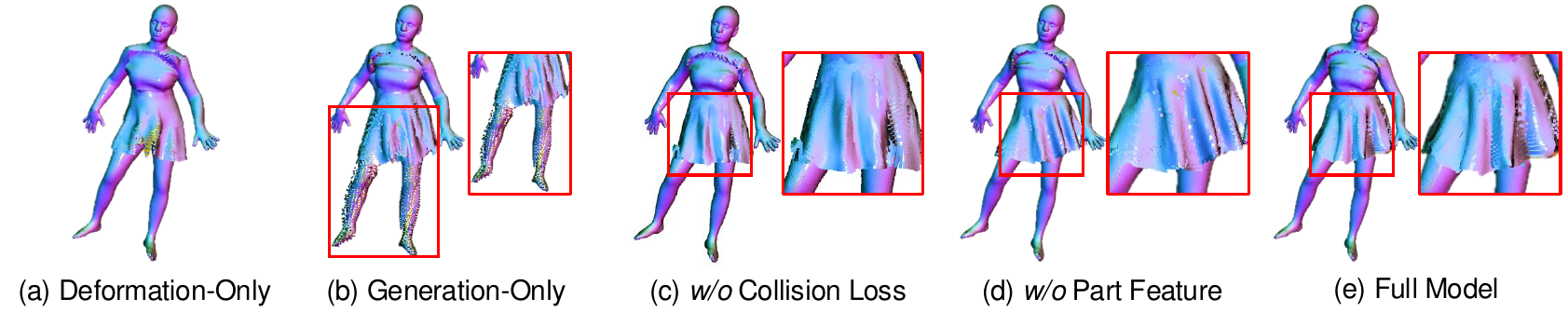}
  \vspace{-1.5ex}
  \caption{\textbf{Ablation study.} (a) shows the LBS-based deformed point cloud, while (b) illustrates the outcomes achieved by only applying free-form generation to the lower body. Ablation (c) examines the effectiveness of the proposed collision loss. Without part-aware pose feature extraction, (d) shows an improper skirt orientation that fails to align with the given pose. Ultimately, our full model (e) showcases the highest visual quality. Please zoom in to examine the details of the generated skirt in the red box. }
  \label{fig:ablation}
  \vspace{-2.ex}
\end{figure*}

\begin{figure}[t]
  \centering
  \includegraphics[width=0.9\linewidth]{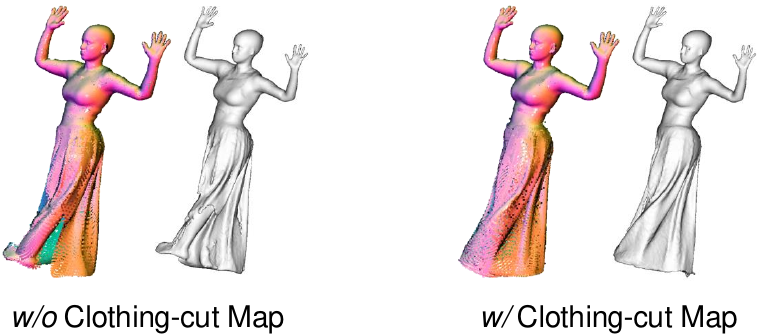}
  \vspace{-1.25ex}
  \caption{\textbf{Ablation study of utilizing clothing-cut maps.} The clothing-cut map effectively guides the free-form generator to model loose garments with a continuous and detailed surface. In contrast, a naive approach to hybrid modeling of the dress causes it to tear apart and the underlying leg to intersect with the garment.}
  \label{fig:ablation_dress}
  \vspace{-2.ex}
\end{figure}

To enhance clarity and better highlight the quality of loose clothing, we specifically present a visual comparison of the generated cloth in \cref{fig:open_surface}.
FITE~\cite{lin2022fite} deforms a learnable implicit template represented in signed distance fields, which struggle with handling the open surfaces of loose skirts. In contrast, our model avoids the surface ``sealing'' issues seen in FITE~\cite{lin2022fite} and eliminates the ``split-up'' artifacts found in POP~\cite{ma2021pop} and SkiRT~\cite{ma2022skirt}, preserving high-quality generation of loose clothing.

\subsection{Ablation Study}
\label{subsec:ablation}
\noindent\textbf{Hybrid Paradigm.} To validate the efficacy of our hybrid paradigm, we implement a simple deformation-only baseline (\ie CloSET~\cite{zhang2023closet} without explicit template decomposition). As shown in \cref{fig:ablation} (a), it still suffers from pant-like artifacts. However, entirely discarding the use of LBS poses challenges in accurately modeling articulated humans as shown in ablation (b), where we generate full points on the lower part of the body from a global pose feature. The results appear noisy and discontinuous, particularly in articulated regions such as legs. This motivates us to take a hybrid approach, which integrates the deformer and the generator modules. \cref{fig:ablation} (e) verifies our analysis that the hybrid method further pushes the upper bound of the expressiveness of LBS-based methods while reasoning about the articulated motion correctly.

\vspace{1.4em}
\noindent\textbf{Collision Loss.} As depicted in \cref{fig:ablation} (c) and (e), the collision loss imposes constraints on the free-form generator to produce loose components that do not intersect with the human body. Overall, we conclude that combining collision loss with pose augmentation yields more robust results.

\vspace{1.4em}
\noindent\textbf{Part-aware Generator.} As depicted in \cref{fig:ablation} (d), the lack of part-aware pose feature encoding results in a misalignment between the skirt’s orientation and the driven pose. This highlights that our structure-aware design facilitates the generator to learn pose conditioning more accurately.

\vspace{1.4em}
\noindent\textbf{Clothing-cut Map.} To assess the efficacy of the clothing-cut map, we conduct an ablation study by removing segmentation guidance and reverting to the default UV map.  In this setting, body points in loose regions are also deformed, leading to the blending of loose clothing points from two branches. As depicted in \cref{fig:ablation_dress}, this results in penetration artifacts, discontinuities, and missing details in the long dress. Note the dress splits around the left leg. In contrast, by introducing clothing-cut maps, the generator can model the whole dress holistically, avoiding conflicts with the deformation module and improving visual quality greatly. 

\section{Conclusion}

We present FreeCloth, a hybrid point-based solution for modeling challenging clothed humans, which integrates LBS deformation and free-form generation to tackle different clothing regions. To synergize the strengths of these two modules, we propose to segment the body surface into unclothed, deformed, and generated regions, yielding a clothing-cut map. Our innovative framework effectively eliminates the pant-like and inhomogeneous density artifacts in prior methods when modeling skirts and long dresses. The free-form generator provides enhanced topological flexibility and expressiveness, enabling our model to generate realistic and high-quality wrinkle details. We assess our model with varying skirt lengths, tightness, and styles, and the experimental results demonstrate the superior representational power of the proposed framework. We believe that this novel hybrid modeling opens up new possibilities in this domain. Additionally, our point-based hybrid modeling can be integrated with recent advancements in 3DGS~\cite{3dgs} to enhance texture rendering, which we plan to explore in future work.

\section*{Acknowledgment}
This work was supported by 2022ZD0114900 and NSFC-6247070125.

{
    \small
    \bibliographystyle{ieeenat_fullname}
    \bibliography{main}
}

\cleardoublepage
\setcounter{page}{1}
\renewcommand\thesection{\Alph{section}}  
\renewcommand\thefigure{\Alph{section}\arabic{figure}} 
\renewcommand\thetable{\Alph{section}\arabic{table}}
\renewcommand\theequation{\Alph{section}\arabic{equation}}
\setcounter{section}{0}
\setcounter{figure}{0}
\setcounter{table}{0}
\setcounter{equation}{0}

\maketitlesupplementary

In Sec.~\ref{supp:sec:imp}, we elaborate on the implementation details of our proposed method and the experimental setups. We provide additional results and extended discussions in Sec.~\ref{supp:sec:results}.

\section{Implementation Details} \label{supp:sec:imp}
\subsection{Model Architecture} \label{supp:sec:supp_arch}

In the implementation of our pose encoder network $\mathcal{E}_d$,
the PointNet++~\cite{qi2017pointnet++} abstracts the point features for $L = 4$ levels, and the numbers of the abstracted points are $2048$, $512$, $128$, and $32$ at each level, respectively. Both the local and global pose codes share a feature dimensionality of $M_p = 256$, whereas the garment code is represented by a $M_g = 64$-dimensional learnable parameter.

The structure-aware pose encoder $\mathcal{E}_g$ for extracting pose feature embedding for the free-form generation module possesses a similar architecture with $\mathcal{E}_d$. Given our focus on modeling skirts and long dresses, we selectively sample posed points from $\boldsymbol{K}_b = 4$ local parts situated on the legs, including the left upper leg, left lower leg, right upper leg, and right lower leg. Although the short skirt doesn’t directly cover the lower legs, their pose still indirectly affects the skirt's movement. Specifically, we uniformly sample $2048$ points from each part, which are then inputted into $\mathcal{E}_g$ to derive part-aware local features. A final global max-pooling layer is prepended to extract the global pose features.

As for the free-form generator $\mathcal{G}$, We modify a simple yet effective style-based point generator, SpareNet~\cite{xie2021style}. SpareNet employs point morphing techniques to map a unit square $[0,1]^2$ onto a 3D surface. Specifically, we utilize $K$ surface elements ($8$ in our experiments) to construct the loose garment. For simplicity, we omit the refiner module and adversarial rendering. Empirically, we observe that refinement following the hybrid modeling of the garment doesn't yield performance improvements. The number of generated points, denoted as $N_g$, is manually configured to either $32768$ for long dresses or $16384$ for skirts.

\subsection{Garment-specific Clothing-cut Map}

Here we provide comprehensive details on computing the garment-specific clothing-cut maps, as outlined in the main paper. Following the methodology in POP~\cite{ma2021pop}, all baseline approaches~\cite{ma2022skirt, lin2022fite, zhang2023closet} uniformly sample point sets from the UV map at a resolution of $256 \times 256$. Specifically, $N_d = 47911$ points are sampled. We start by segmenting the unclothed regions, including the head, hands, and feet, which contain up to $N_u = 13240$ points. 

\begin{figure*}[h]
  \centering
  \includegraphics[width=0.85\linewidth]{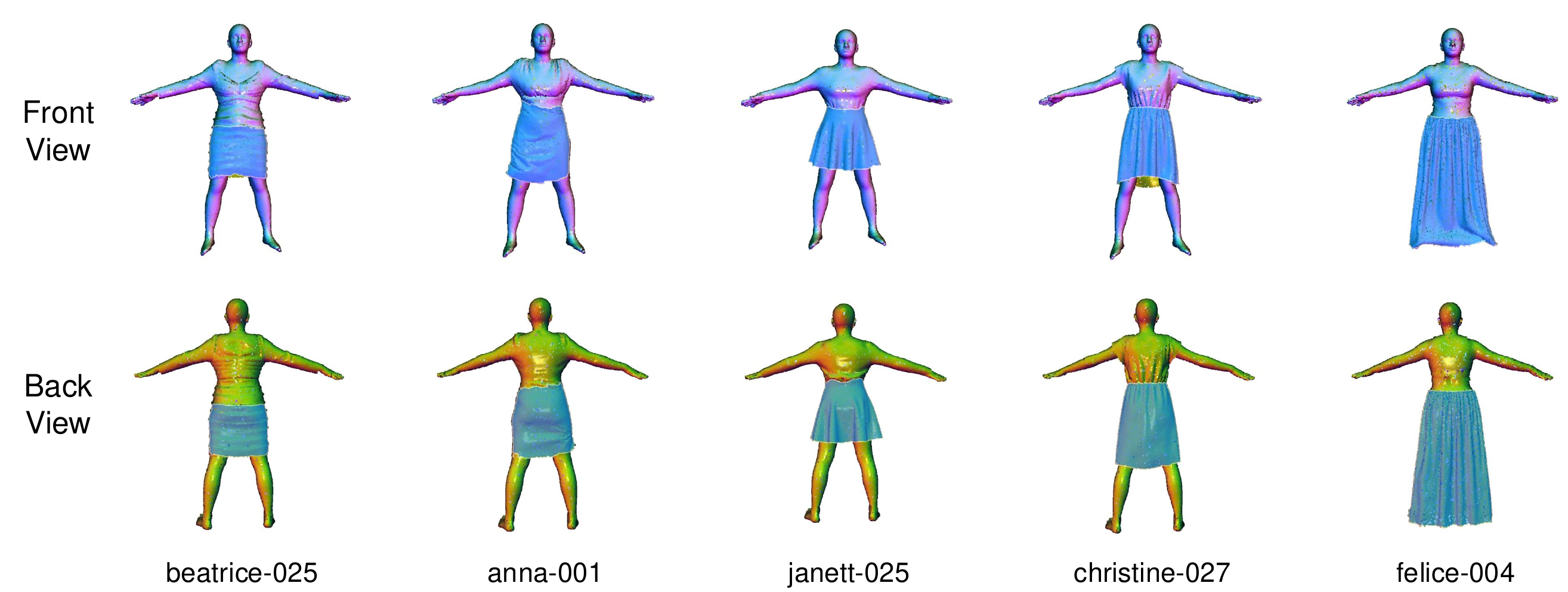}
  \vspace{-1.5ex}
  \caption{\textbf{The segmented loose regions of each cloth in the ReSynth~\cite{ma2021pop} dataset.} We identify the loose regions in the front and back view normal maps utilizing the segmentation model SAM~\cite{kirillov2023sam}.}
  \label{fig:clothing_seg}
\end{figure*}

\begin{figure*}[h]
  \centering
  \includegraphics[width=0.72\linewidth]{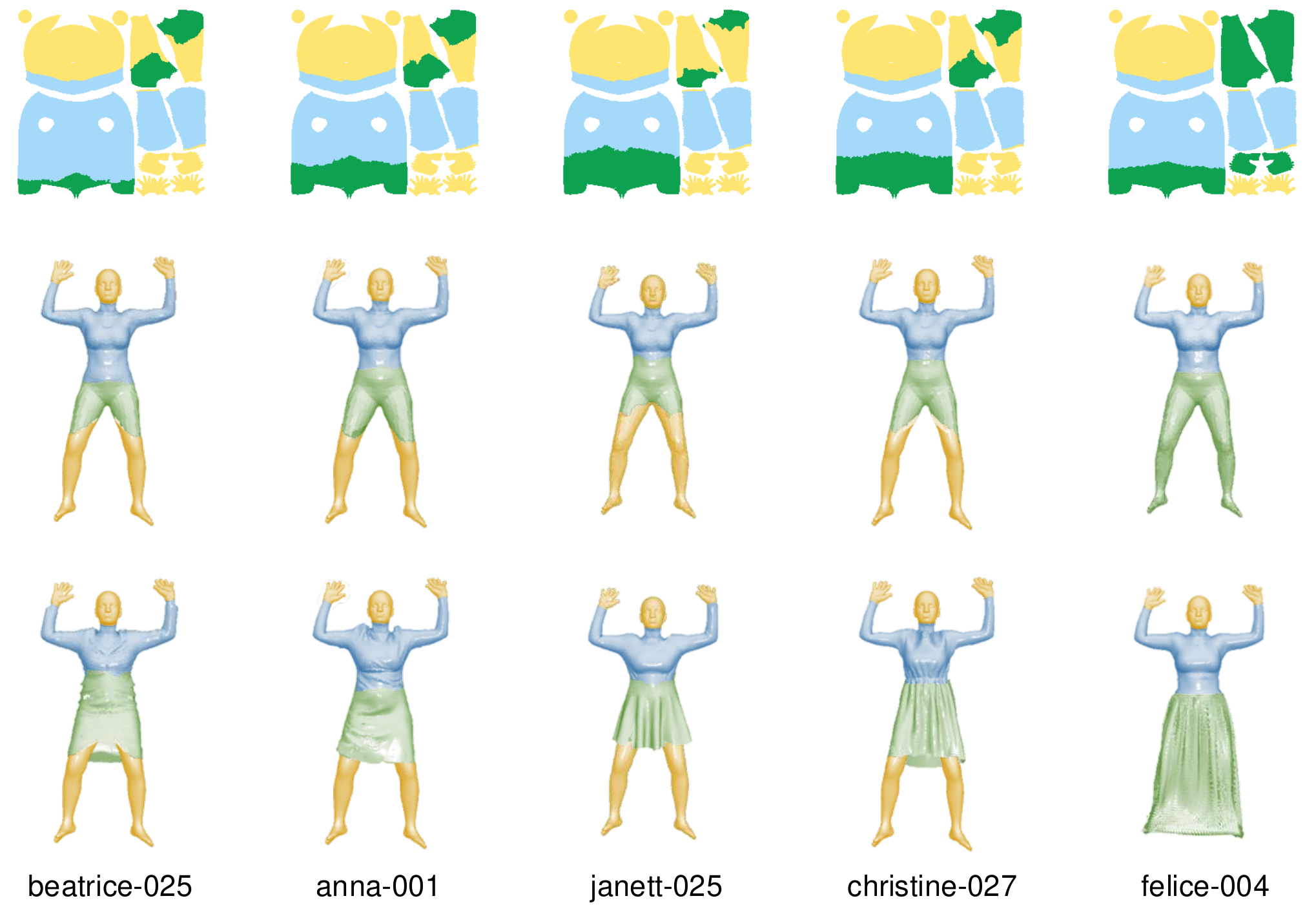}
  \vspace{-2ex}
  \caption{\textbf{The clothing-cut maps for five subjects in the ReSynth~\cite{ma2021pop} dataset.} The first row depicts the clothing-cut maps distinguished by three different colors, while the second row illustrates the corresponding segmented regions. Specifically, the \textcolor{myyellow}{yellow} color denotes the masked region, the \textcolor{myblue}{blue} indicates the body parts requiring deformation, and the \textcolor{mygreen}{green} marks the loose parts to be modeled utilizing free-form generation. Finally, the last row displays the complete predictions generated by our model.}
  \label{fig:clothing_map}
\end{figure*}

Then we apply the off-the-shelf image segmentation model, SAM~\cite{kirillov2023sam}, to automatically identify the loose region. Specifically, we select the frame that closely resembles the canonical pose in the training sequence and render the front and back view normal maps to cover all body points. These normal maps are fed into SAM to locate loose clothing including skirts and dresses. The segmented results are shown in \cref{fig:clothing_seg}.  We back-project the detected pixel coordinates into 3D space and employ nearest neighbor search to assign each point on the UV map to the full scan, filtering the corresponding loose parts on the body surface. The extracted clothing-cut maps for all 5 subjects from the ReSynth~\cite{ma2021pop} dataset are visualized in \cref{fig:clothing_map}. 

To ensure fair comparisons, we merge points from three sources, \ie combining $N_u$, $N_d$, and $N_g$ points, and employ farthest point sampling (FPS) to obtain the final full point set with $N=47911$ points to match the baselines~\cite{ma2021pop, ma2022skirt}. 

\subsection{Training}
We train our network for $1000$ epochs on the ReSynth~\cite{ma2021pop} dataset, using the Adam~\cite{kingma2014adam} optimizer with a batch size of $8$ and a learning rate of $3.0\times {10}^{-4}$. The loss weights are set to $\lambda_p = 1\times {10}^4$, $\lambda_n = 1.0$, $\lambda_{rd} = 2 \times 10^3$, $\lambda_{rg} = 1$ and $\lambda_{col} = 2\times10^{-2}$ to balance loss terms. Following previous works~\cite{ma2021pop, zhang2023closet}, we only activate the normal loss from the $400^{th}$ epoch. The training procedure takes about 20 hours on a single RTX 3090 GPU. Given limited 3D training data, we enhance the robustness of our free-form generator to out-of-distribution poses by balancing the pose distribution. Specifically, we apply random horizontal flips along the $x$-axis, leveraging the symmetry of the human body.

\subsection{Baselines}
\label{supp:sec:supp_baseline}
For POP~\cite{ma2021pop} and FITE~\cite{lin2022fite}, we directly utilize the official model weight provided for inference. As for SkiRT~\cite{ma2022skirt}, we train the model using the official code and successfully reproduce the results reported in the original paper. We perform inference using the trained model weight.

\begin{figure}[h]
  \centering
  \includegraphics[width=\linewidth]{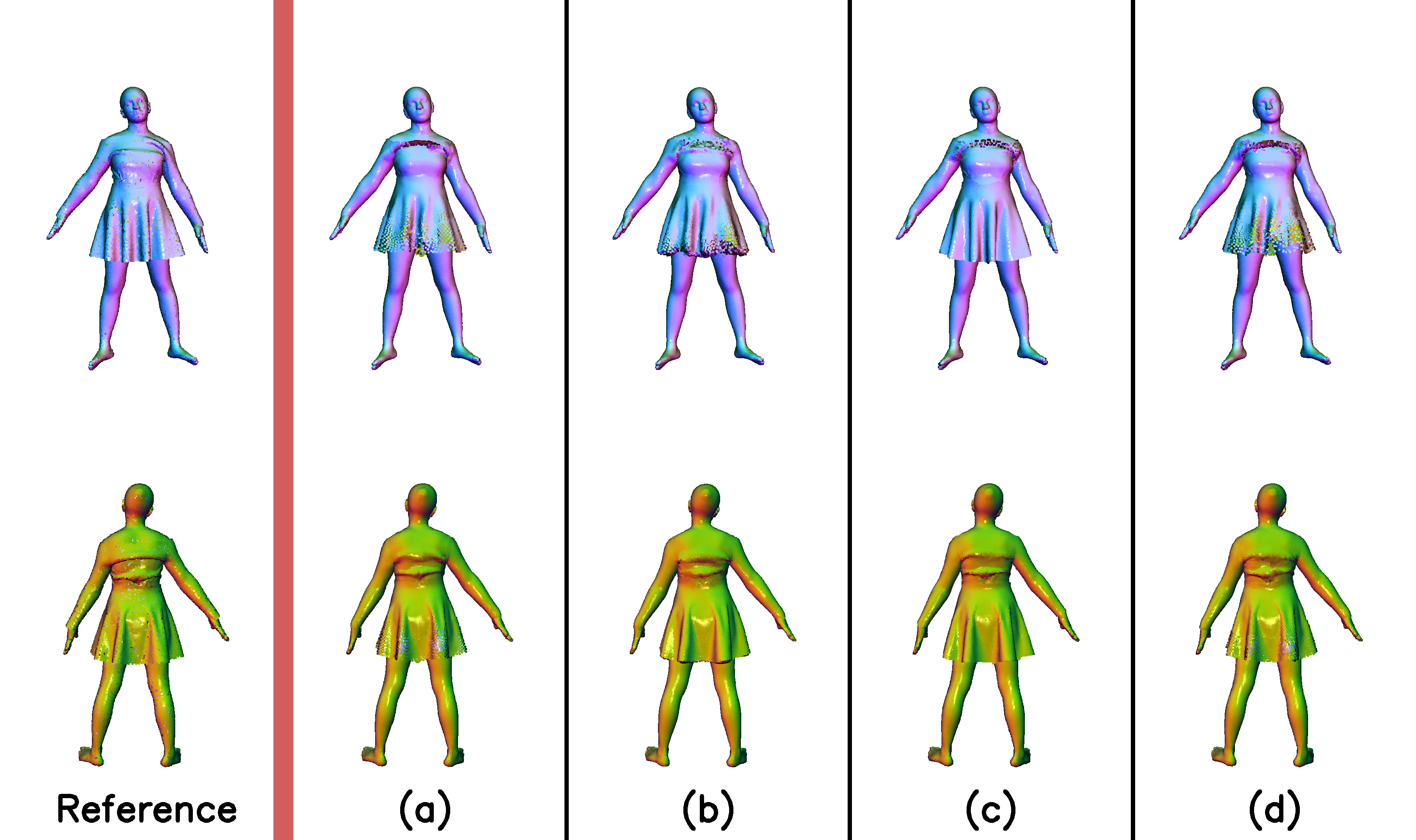}
  \vspace{-3.5ex}
  \caption{\textbf{Example of perceptual study image.} We randomize the ordering of the results of different methods per example. We always put the GT result in the leftmost column. }
  \label{fig:interface}
\end{figure}

\subsection{Details on Perceptual Study} \label{supp:sec:supp_perceptual}
We follow the official rendering scripts including camera and lighting configurations for baseline methods~\cite{ma2021pop,ma2022skirt, lin2022fite} and ours, where the output point cloud is rendered using a surfel-based renderer in Open3D~\cite{zhou2018open3d} with a point size of 5. To assess the geometric visual quality, we render the front and the back views at a high resolution of $1024\times 1024$. The deployed baseline models are discussed above in \cref{supp:sec:supp_baseline}. 

50 participants are presented with a set of 25 examples consisting of different subjects and poses, \textbf{randomly} sampled from the ReSynth~\cite{ma2021pop} dataset results without cherry-picking. In each example, the GT reference is always put in the leftmost column, and we randomize the ordering of the results of different methods on the right. \cref{fig:interface} shows an example. For each example, the participants are asked to select the most preferred single option based on the following two criteria: (1) realism, wrinkle details, smoothness, uniformity, and the presence of artifacts contribute to the overall visual quality of the clothing shape; (2) the similarity to the reference GT result. Due to the inherent randomness in the generated results, an exact match with the reference effect may not be necessary. Therefore, priority should be given to the first point, which is the overall visual quality.

\section{Extended Results and Discussions} \label{supp:sec:results}

\subsection{Discussions on the Evaluation Metric} \label{supp:sec:supp_metric}
\begin{figure}[h]
  \centering
  \includegraphics[width=0.63\linewidth]{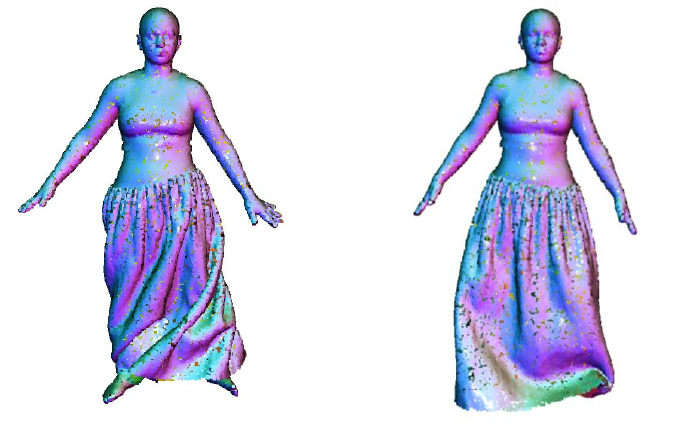}
  \vspace{-1.5ex}
  \caption{\textbf{An example illustrating the stochasticity of clothing shape with two similar given poses.} }
  \label{fig:supp_random}
\end{figure}

As pointed out by DPF~\cite{prokudin2023dpf} and FITE~\cite{lin2022fite}, we emphasize that conventional metrics used in the previous works~\cite{ma2021pop,ma2022skirt,saito2021scanimate}, Chamfer distance (CD) and $\mathcal{L}_1$ normal discrepancy (NML) implicitly assume a one-to-one mapping from body pose to the clothing shape. However, in reality, the clothing shape possesses diversity and randomness which can be influenced by many other factors such as the motion speed and the history~\cite{prokudin2023dpf}. Consequently, given a similar or same pose, multiple clothing statuses can be reasonable, as illustrated in \cref{fig:supp_random}. Our model generates plausibly-looking results that, may not conform strictly to the ground truth, hence obtaining high CD errors.

\begin{figure}[h]
  \centering
  \includegraphics[width=0.9\linewidth]{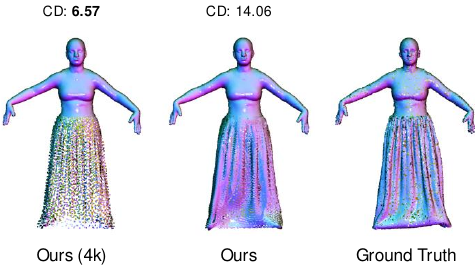}
  \vspace{-1ex}
  \caption{\textbf{Illustration of the paradox of lower CD error with worse visual quality.} Reducing the number of generated points signficantly decreases the CD error, yet results in visually unsatisfactory outcomes with non-uniform point density.}
  \label{fig:supp_cd}
\end{figure}

To further highlight the limitations of the CD metric, we examine a case involving generated points for a long dress. When reducing the number of points generated by the free-form generator $N_g$ from $32768$ to $4096$, the CD error substantially decreases (from $14.06$ to $6.57$, a $53.3\%$ reduction), as shown in \cref{fig:supp_cd}. However, this reduction comes at the cost of point density uniformity and detail, such as wrinkles. This observation has been proven in previous works that the CD metric lacks awareness of the point density distribution~\cite{wu2021density, huang2022learning, liu2020morphing}. This phenomenon also helps explain why methods like POP~\cite{ma2021pop} and SkiRT~\cite{ma2022skirt} achieve lower CD errors despite significantly lower point densities on loose skirts and dresses.

Building on the limitations discussed, we employ a generation-based metric, FID, which compares distributions and relaxes the strict one-to-one mapping constraint, alongside the reconstruction-based MSE loss to assess our model’s quality holistically. This evaluation approach better aligns with the model’s objective.

\begin{table*}[t]
\centering
\caption{\textbf{Perceptual study results on the ReSynth~\cite{ma2021pop} dataset for each subject.} We report the preference rates (PR-H) obtained from a perceptual study involving 50 participants, alongside the preference rates (PR-G) voted by the GPT-4o~\cite{achiam2023gpt4} model. The final scores are generally consistent with those of the human participants. The best results are highlighted in \textbf{bold}, and the second best are \underline{underlined}. The subject IDs are listed in descending order based on the looseness of the clothing. }
\label{tab:user_study_subject}
\setlength{\tabcolsep}{3pt}
\begin{tabular}{@{}ccccccccccccc@{}}
\toprule
Subject ID                     & \multicolumn{2}{c}{felice-004} & \multicolumn{2}{c}{janett-025}  & \multicolumn{2}{c}{christine-027} & \multicolumn{2}{c}{anna-001}     & \multicolumn{2}{c}{beatrice-025} & \multicolumn{2}{c}{Average} \\ \midrule
\multicolumn{1}{c|}{Method}    & PR-H~$\uparrow$   & \multicolumn{1}{c|}{PR-G~$\uparrow$} & PR-H~$\uparrow$   & \multicolumn{1}{c|}{PR-G~$\uparrow$} & PR-H~$\uparrow$   & \multicolumn{1}{c|}{PR-G~$\uparrow$} & PR-H~$\uparrow$    & \multicolumn{1}{c|}{PR-G~$\uparrow$}  & PR-H~$\uparrow$  & \multicolumn{1}{c|}{PR-G~$\uparrow$}           & PR-H~$\uparrow$ & PR-G~$\uparrow$ \\ \midrule
\multicolumn{1}{c|}{POP~\cite{ma2021pop}}  & 0.4\% & \multicolumn{1}{c|}{0.0\%}  & 0.8\%  & \multicolumn{1}{c|}{0.0\%} & 0.8\%  & \multicolumn{1}{c|}{0.0\%} & 1.2\%  & \multicolumn{1}{c|}{0.0\%} & 0.8\% & \multicolumn{1}{c|}{0.0\%} & 0.8\% & 0.0\% \\ 
\multicolumn{1}{c|}{SkiRT~\cite{ma2022skirt}} & 3.2\% & \multicolumn{1}{c|}{0.0\%} & 0.0\%  & \multicolumn{1}{c|}{0.0\%} & 3.2\%  & \multicolumn{1}{c|}{0.0\%} & 12.8\% & \multicolumn{1}{c|}{\underline{20.0\%}} & 10.4\% & \multicolumn{1}{c|}{\textbf{60.0\%}} & 5.9\% & 16.0\% \\ 

\multicolumn{1}{c|}{FITE~\cite{lin2022fite}} & \underline{28.0\%} & \multicolumn{1}{c|}{\underline{40.0\%}} & \underline{6.0\%} & \multicolumn{1}{c|}{\underline{20.0\%}} & \underline{10.8\%} & \multicolumn{1}{c|}{\underline{20.0\%}} & \textbf{54.0\%} & \multicolumn{1}{c|}{\textbf{40.0\%}} & \textbf{50.8\%} & \multicolumn{1}{c|}{\underline{20.0\%}} & \underline{29.9\%} & \underline{28.0\%} \\ 

\multicolumn{1}{c|}{Ours} & \textbf{68.4\%}  & \multicolumn{1}{c|}{\textbf{60.0\%}} & \textbf{93.2\%} &  \multicolumn{1}{c|}{\textbf{80.0\%}} & \textbf{85.2\%} & \multicolumn{1}{c|}{\textbf{80.0\%}} & \underline{32.0\%} &  \multicolumn{1}{c|}{\textbf{40.0\%}} & \underline{38.0\%} & \multicolumn{1}{c|}{\underline{20.0\%}} & \textbf{63.4\%} & \textbf{56.0\%} \\ 
\bottomrule
\end{tabular}
\end{table*}

\normalsize %
\begin{table*}[h]
\centering
\caption{\textbf{Additional quantitative comparison of different methods on the ReSynth~\cite{ma2021pop} dataset for each subject.}}
\label{tab:resynth_per}
\setlength{\tabcolsep}{6pt}
\begin{tabular}{@{}ccccccccccccc@{}}
\toprule
Subject ID                     & \multicolumn{2}{c}{felice-004} & \multicolumn{2}{c}{christine-027}  & \multicolumn{2}{c}{janett-025} & \multicolumn{2}{c}{anna-001}     & \multicolumn{2}{c}{beatrice-025} \\ \midrule
\multicolumn{1}{c|}{Method}    & CD~$\downarrow$   & \multicolumn{1}{c|}{NML~$\downarrow$} & CD~$\downarrow$   & \multicolumn{1}{c|}{NML~$\downarrow$} & CD~$\downarrow$   & \multicolumn{1}{c|}{NML~$\downarrow$} & CD~$\downarrow$    & \multicolumn{1}{c|}{NML~$\downarrow$}  & CD~$\downarrow$             & NML~$\downarrow$           \\ \midrule
\multicolumn{1}{c|}{POP~\cite{ma2021pop}}       & 
7.34  & \multicolumn{1}{c|}{1.24}  & 1.72 & \multicolumn{1}{c|}{0.97} & 
1.24  & \multicolumn{1}{c|}{0.89} & 0.62 & \multicolumn{1}{c|}{0.82} & 0.34 & 0.75          \\
\multicolumn{1}{c|}{SkiRT~\cite{ma2022skirt}} & 
6.45 & \multicolumn{1}{c|}{1.25} & 1.54 & \multicolumn{1}{c|}{0.99} & 
1.10  & \multicolumn{1}{c|}{0.82} & 0.58 & \multicolumn{1}{c|}{0.81} & 0.31 & 0.77          \\
\multicolumn{1}{c|}{FITE~\cite{lin2022fite}} & 11.27
 & \multicolumn{1}{c|}{2.38} & 2.16  & \multicolumn{1}{c|}{1.15} & 1.52
  & \multicolumn{1}{c|}{1.05} & 0.74 & \multicolumn{1}{c|}{0.91} & 0.46 & 0.85  \\
\multicolumn{1}{c|}{Ours} & 10.61  & \multicolumn{1}{c|}{1.78} & 2.18  &  \multicolumn{1}{c|}{1.01} & 1.59   & \multicolumn{1}{c|}{0.94}  & 0.81 &  \multicolumn{1}{c|}{0.84} & 0.48 & 0.74  \\
\bottomrule
\end{tabular}
\end{table*}

\begin{figure*}[!t]
  \centering
  \includegraphics[width=1\linewidth]{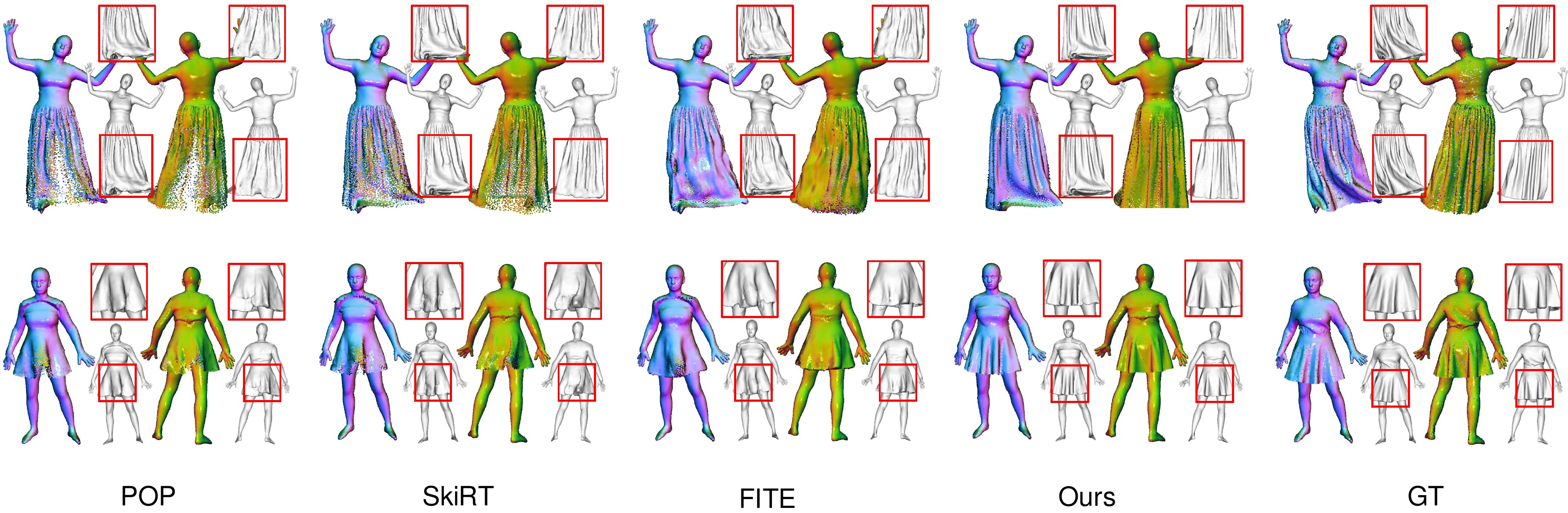}
  \vspace{-4ex}
  \caption{\textbf{Qualitative comparison between baselines and our model for modeling loose clothing, with highlighted details.} Subject IDs from top to bottom: ``felice-004'' and ``janett-025''. Best viewed zoomed-in on a color screen.}
  \label{fig:supp_sota_more}
\end{figure*}

\subsection{Quantitative Results}

In addition to the user study, we also employ the GPT-4o model~\cite{achiam2023gpt4} to select the best result across all methods. The testing prompt is as follows: "\textit{Select the most preferred option based on the following two criteria: (1) realism, wrinkle details, smoothness, uniformity, and the presence of artifacts, which contribute to the overall visual quality of the clothing shape; (2) similarity to the reference ground truth (GT) result. Priority should be given to the first criterion, which emphasizes the overall visual quality.}" The per-subject preference rates of human users and GPT-4o are presented in~\cref{tab:user_study_subject}, denoted as PR-H and PR-G, respectively. As shown, the results are generally consistent between the two, with our method demonstrating significant advantages in handling challenging cases.  For the two most difficult skirts, the preference rates from GPT-4o reach $80\%$, while human users show a preference rate exceeding $85\%$, confirming the effectiveness of our hybrid design in modeling loose clothing. For tighter skirts, our model performs on par with FITE~\cite{lin2022fite} and SkiRT~\cite{ma2022skirt}, both of which rely purely on LBS but still generate promising results.

For reference, we also follow previous works~\cite{ma2021pop, zhang2023closet, lin2022fite, ma2022skirt} to evaluate the Chamfer Distance (CD) and the $\mathcal{L}_1$ normal discrepancy (NML), as specified by the formulas in the main paper. The default units for reporting CD and NML are $\times {10}^{-4} m^2$ and $\times {10}^{-1}$, respectively. \cref{tab:resynth_per} presents the quantitative errors on the ReSynth~\cite{ma2021pop} dataset. Notably, while FITE~\cite{lin2022fite} exhibits the highest visual quality among the baselines, it also results in significantly larger quantitative errors. This further verifies that the CD metric may not accurately reflect performance, as discussed in Sec.~\ref{supp:sec:supp_metric}. Our model shows comparable performance to FITE in CD errors while significantly reducing the normal discrepancy, which corroborates our observation that FITE generates unnatural and excessively bent wrinkles, whereas our model effectively captures complex local details.

\begin{figure}[t]
  \centering
  \includegraphics[width=0.95\linewidth]{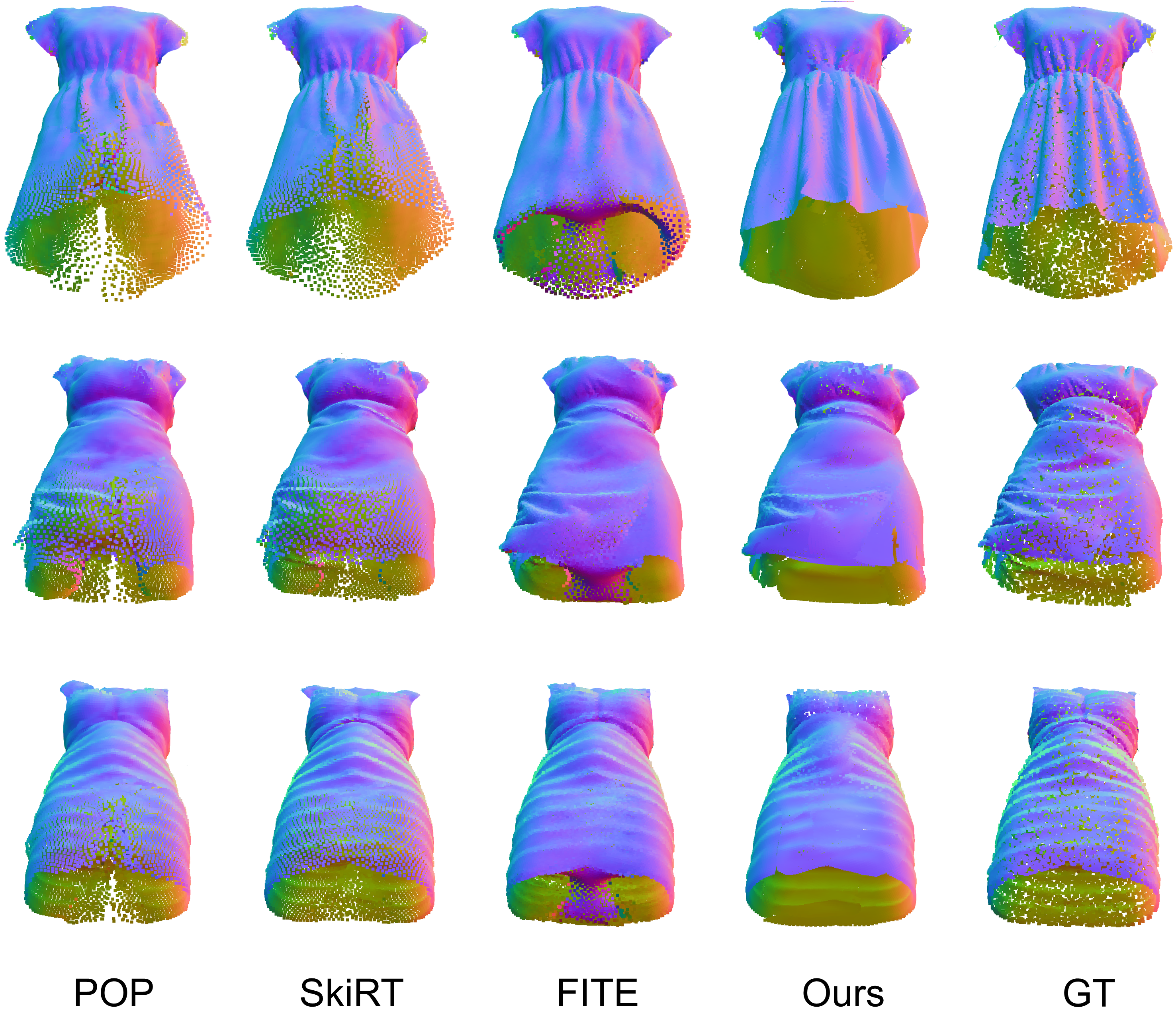}
  \vspace{-1ex}
  \caption{\textbf{Visualization results of loose clothing.} As shown, while FITE~\cite{lin2022fite} successfully captures intricate details like wrinkles in tighter skirts, it still faces the ``open-surface'' challenge. In contrast, our model generates more accurate geometry and achieves superior visual quality.}
  \label{fig:open_surface_supp}
\end{figure}

\subsection{More Qualitative Results}
\begin{figure*}
  \centering
  \includegraphics[width=\linewidth]{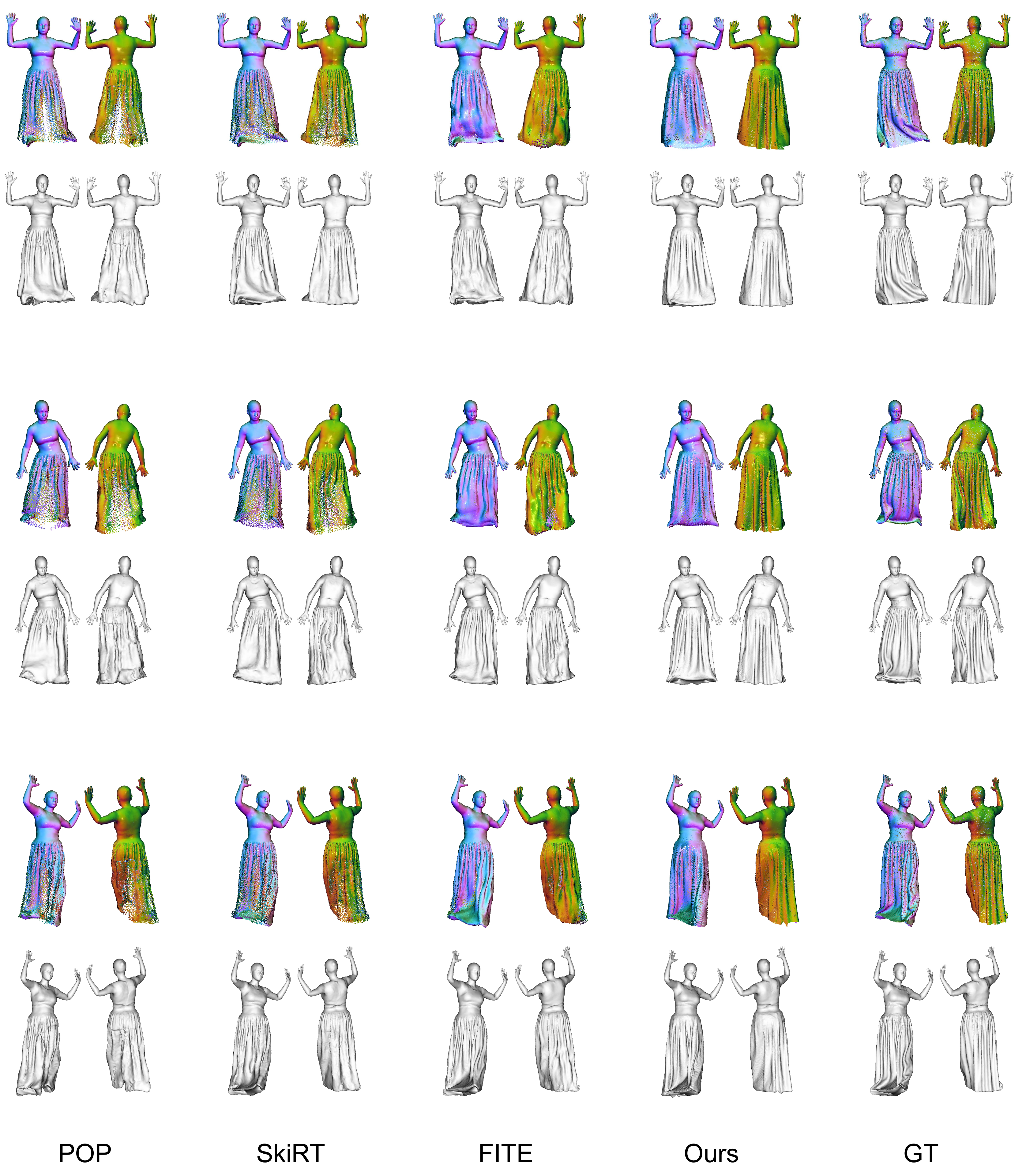}
  \caption{\textbf{Additional Qualitative comparison between baselines and our model.} The subject ID is ``felice-004'' from the ReSynth~\cite{ma2021pop} dataset. Best viewed zoomed-in on a color screen.}
  \label{fig:supp_sota_felice}
\end{figure*}

\begin{figure*}
  \centering
  \includegraphics[width=\linewidth]{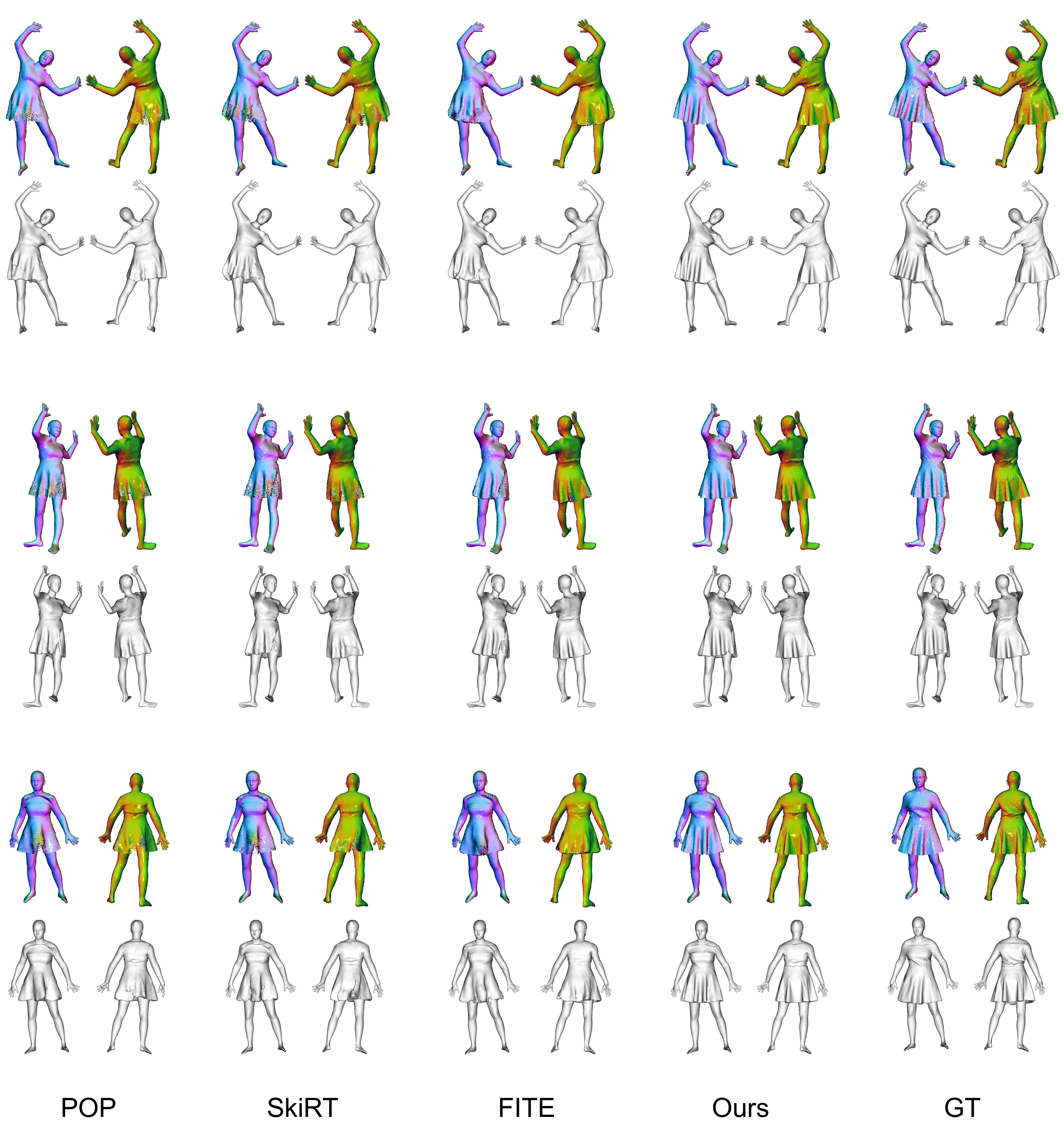}
  \caption{\textbf{Additional Qualitative comparison between baselines and our model.} The subject ID is ``janett-025'' from the ReSynth~\cite{ma2021pop} dataset. Best viewed zoomed-in on a color screen.}
  \label{fig:supp_sota_janett}
\end{figure*}

\begin{figure*}
  \centering
  \includegraphics[width=\linewidth]{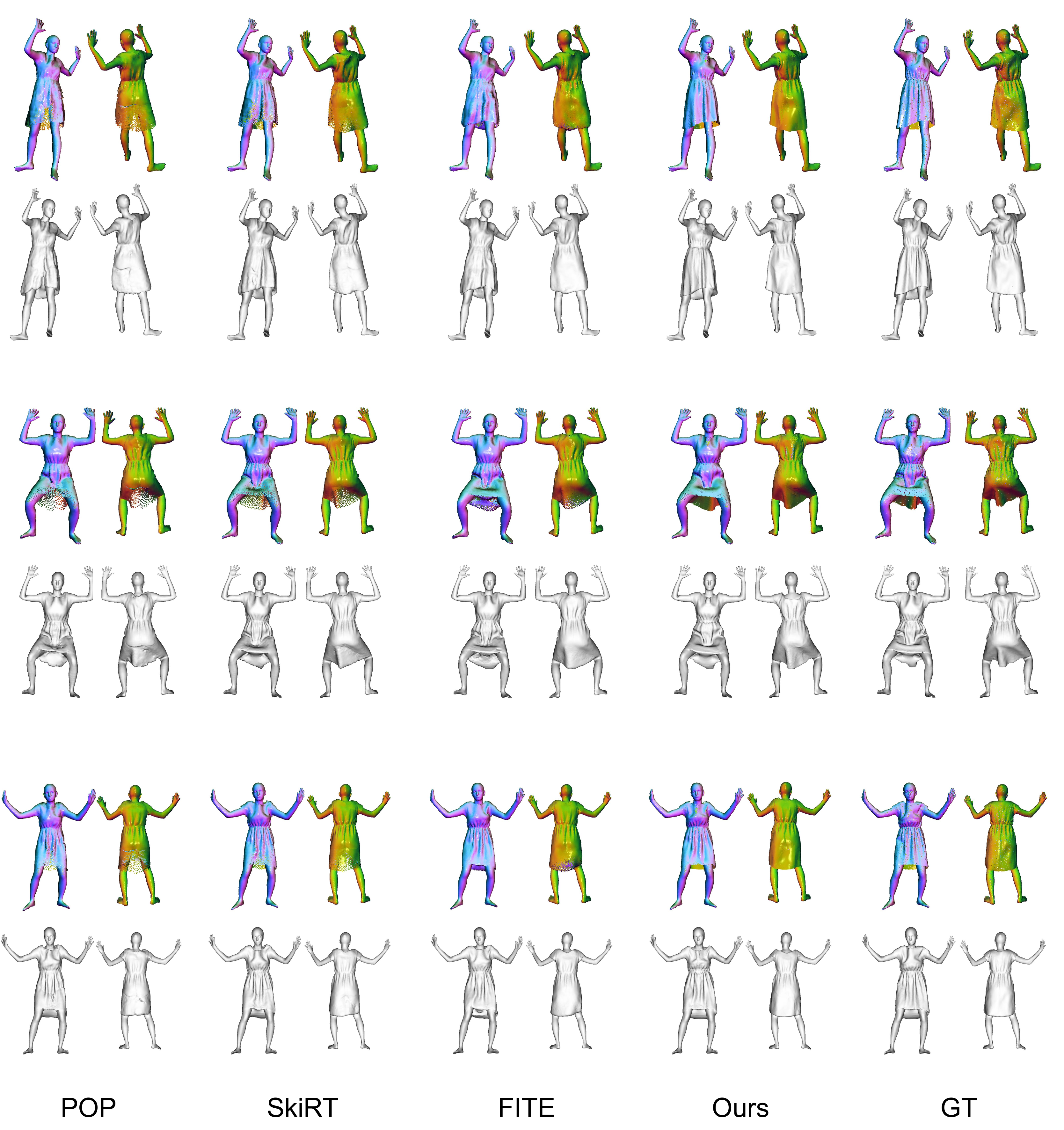}
  \caption{\textbf{Additional Qualitative comparison between baselines and our model.} The subject ID is ``christine-027'' from the ReSynth~\cite{ma2021pop} dataset. Best viewed zoomed-in on a color screen.}
  \label{fig:supp_sota_christine}
\end{figure*}

\begin{figure*}
  \centering
  \includegraphics[width=\linewidth]{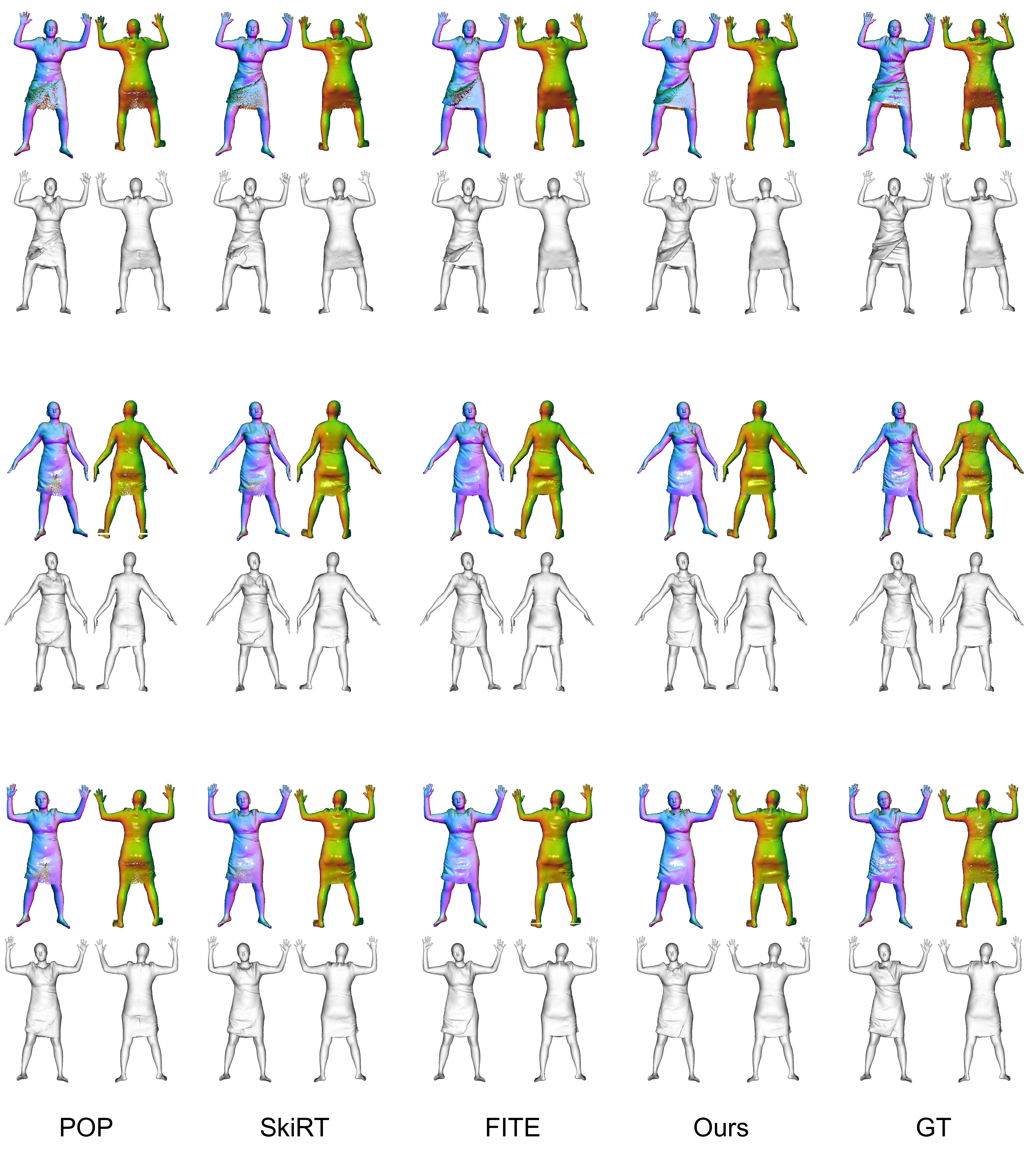}
  \caption{\textbf{Additional Qualitative comparison between baselines and our model.} The subject ID is ``anna-001'' from the ReSynth~\cite{ma2021pop} dataset. Best viewed zoomed-in on a color screen.}
  \label{fig:supp_sota_anna}
\end{figure*}

\begin{figure*}
  \centering
  \includegraphics[width=\linewidth]{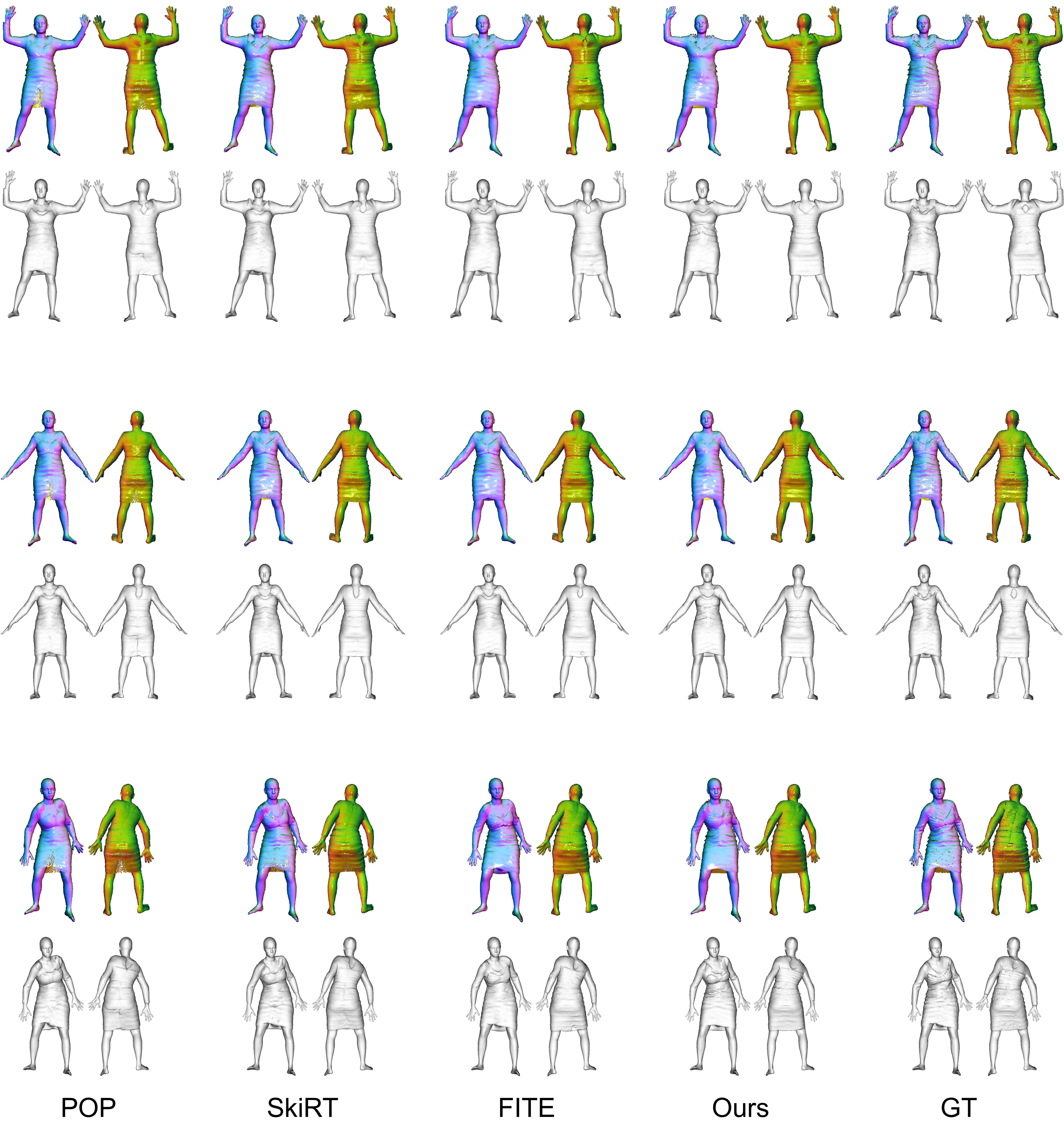}
  \caption{\textbf{Additional Qualitative comparison between baselines and our model.} The subject ID is ``beatrice-025'' from the ReSynth~\cite{ma2021pop} dataset. Best viewed zoomed-in on a color screen.}
  \label{fig:supp_sota_beatrice}
\end{figure*}

In this section, we present additional visualization comparisons that extend the results discussed in the main paper. \cref{fig:supp_sota_more} illustrates the details of the generated loose clothing, which are highlighted within the red box. Furthermore, we showcase three testing examples for each of the five subjects from the ReSynth~\cite{ma2021pop} dataset, as illustrated in \cref{fig:supp_sota_felice,fig:supp_sota_janett,fig:supp_sota_christine,fig:supp_sota_anna,fig:supp_sota_beatrice}. We recommend zooming in to observe finer details, particularly the wrinkles in skirts and dresses. Please refer to the visualization demo in our supplementary materials, which includes sequences of testing data to better demonstrate the high-quality performance of our method. 

As discussed in the main paper, the advantages of our hybrid modeling approach become particularly evident with loose skirts or dresses, as illustrated by the examples in \cref{fig:supp_sota_felice,fig:supp_sota_janett}. For tighter skirts, LBS-based models like FITE~\cite{lin2022fite} already perform well since the clothing adheres closely to the body. As shown in \cref{fig:supp_sota_anna,fig:supp_sota_beatrice}, FITE~\cite{lin2022fite} generates nearly perfect outputs that closely resemble the ground truth, and our model produces results comparable to those of FITE~\cite{lin2022fite}. However, it is noteworthy that FITE~\cite{lin2022fite} still fails to fully eliminate redundant points on the open surface of tighter skirts (see \cref{fig:open_surface_supp}).

Above all, our approach stands out in its ability to operate without subject-specific templates coupled with LBS fields, allowing for more flexible, multi-subject modeling. This opens up new possibilities for avatar modeling while maintaining high performance.

\subsection{Multi-Subject Experiments}
In this study, we explore the potential of hybrid modeling for loose clothing and significantly improve the performance under a single-subject setting. Nevertheless, our hybrid paradigm can be naturally extended to modeling multiple garments, conditioned on various global garment codes. Experimental results show that our unified, multi-subject model demonstrates promising performance in modeling various types of skirts and long dresses, confirming the expressive power of our free-form generator.

\begin{figure}[t]
  \centering
  \includegraphics[width=1\linewidth]{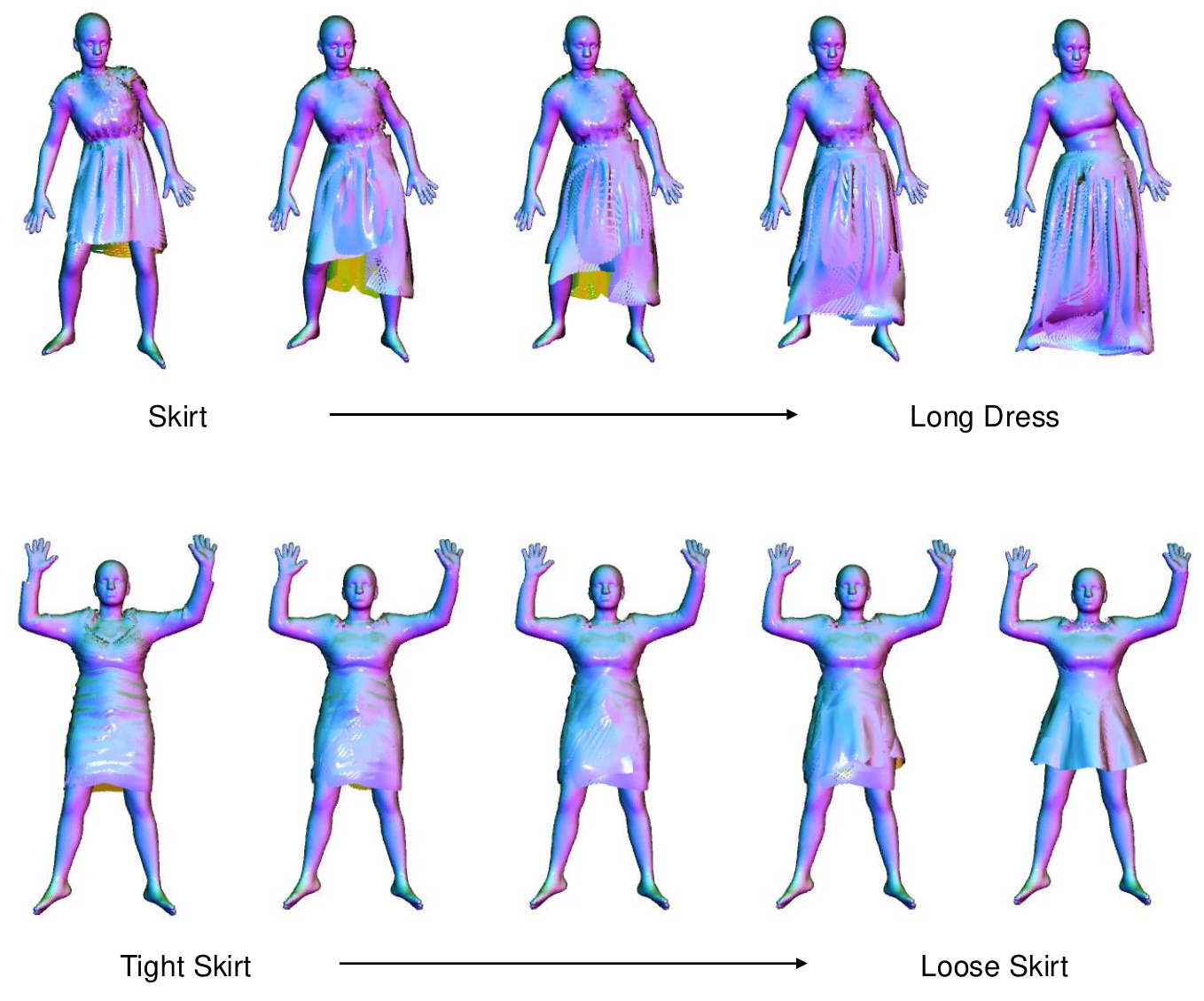}
  \vspace{-4ex}
  \caption{\textbf{Interpolation results when varying the length and the tightness of the skirt.}}
  \label{fig:inter}
\end{figure}

To explore the learned latent space of garment codes, we perform interpolation experiments focusing on two crucial attributes: length and tightness. As shown in \cref{fig:inter}, our model allows effective control over garment length through manipulation of the garment code. Furthermore, when varying the tightness, the generated skirts smoothly transition from tight to loose. In summary, our model successfully disentangles pose-related effects from garment-specific features, providing controllable and realistic generation results.

\subsection{Fitting Non-skirt Clothing}
\begin{figure}[h]
  \centering
  \includegraphics[width=0.5\linewidth]{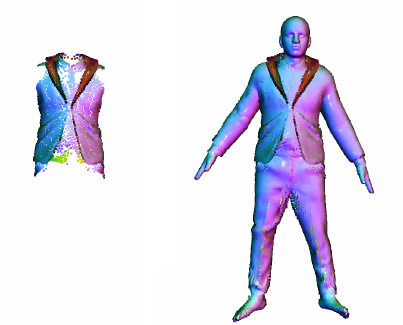}
  \vspace{-1ex}
  \caption{\textbf{Our hybrid model can also handle non-skirt clothing such as suits.} As shown on the left-hand side, the free-form generation module can model loose regions such as collars.}
  \label{fig:non_skirt}
\end{figure}

Although the main focus of this paper is to investigate the hybrid modeling of loose garments such as skirts and long dresses, we also conduct experiments to handle non-skirt clothing, \textit{e.g.} suits. Note that the global pose feature is extracted from the PointNet++~\cite{qi2017pointnet++} without part-aware local feature learning. Visualization results (\cref{fig:non_skirt}) illustrate the generator's capacity to autonomously learn and represent loose components, such as collars. This demonstrates the flexibility and the promising expressiveness of the proposed free-form generator.

\begin{table*}[t]
\centering
\caption{\textbf{Ablation study of the free-form generation module on the ReSynth~\cite{ma2021pop} dataset.} In the setting \textit{Ours$^{\ast}$}, the free-form generator is removed, relying solely on body point deformation to model loose clothing.}
\label{tab:supp_ablate_free}
\resizebox{\linewidth}{!}{ %
\begin{tabular}{@{}c|cc|cccccccccc@{}}
\toprule
Subject                     & \multicolumn{2}{c|}{All} & \multicolumn{2}{c}{felice-004} & \multicolumn{2}{c}{janett-025}  & \multicolumn{2}{c}{christine-027} & \multicolumn{2}{c}{anna-001}     & \multicolumn{2}{c}{beatrice-025} \\ \midrule
\multicolumn{1}{c|}{Metric}    & FID~$\downarrow$   & \multicolumn{1}{c|}{MSE~$\downarrow$} & FID~$\downarrow$   & \multicolumn{1}{c}{MSE~$\downarrow$} & FID~$\downarrow$   & \multicolumn{1}{c}{MSE~$\downarrow$} & FID~$\downarrow$    & \multicolumn{1}{c}{MSE~$\downarrow$}  & FID~$\downarrow$  & \multicolumn{1}{c}{MSE~$\downarrow$}  & FID~$\downarrow$ & MSE~$\downarrow$ \\ \midrule
\multicolumn{1}{c|}{\textbf{Ours$^{\ast}$}} & 56.23 & \multicolumn{1}{c|}{2.73} & 63.12 & \multicolumn{1}{c}{5.72} & 52.10 &  \multicolumn{1}{c}{2.06} & 59.29 & \multicolumn{1}{c}{2.41} & 51.68 &  \multicolumn{1}{c}{\textbf{1.84}} & 54.96 & \textbf{1.62} \\ 
\multicolumn{1}{c|}{\textbf{Ours}} & \textbf{37.75} & \multicolumn{1}{c|}{\textbf{2.61}} & \textbf{42.41}  
& \multicolumn{1}{c}{\textbf{5.24}} & \textbf{27.95} & \multicolumn{1}{c}{\textbf{1.92}} & \textbf{37.43} 
& \multicolumn{1}{c}{\textbf{2.35}} & \textbf{39.63} & \multicolumn{1}{c}{1.89} & \textbf{41.24} & 1.68 \\
\bottomrule
\end{tabular}}
\end{table*}

\subsection{More Ablation Studies}
\begin{figure}[t]
  \centering
  \includegraphics[width=0.7\linewidth]{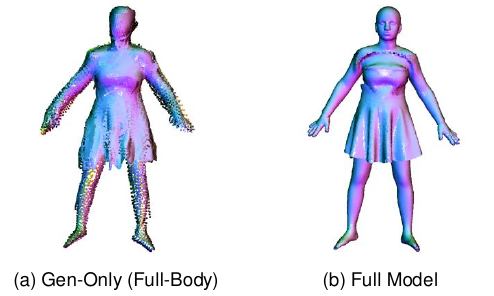}
  \vspace{-1ex}
  \caption{\textbf{Ablation study of the full-body free-form generation.} (a) Completely discarding LBS deformation results in a significant performance drop when compared to (b) our full model.}
  \label{fig:ablate_def}
\end{figure}

\paragraph{Effects of the Hybrid Paradigm.} To evaluate the efficacy of the proposed free-form generator, we quantitatively assess the deformation-only variant using the ReSynth~\cite{ma2021pop} dataset. As shown in \cref{tab:supp_ablate_free}, this variant achieves an average FID of 56.23 and MSE of 2.74, comparable to SkiRT~\cite{ma2022skirt}. In contrast, our full model substantially improves these metrics, demonstrating the effectiveness of the free-form generation module in capturing the dynamics of loose clothing. Additionally, we examine a generation-only variant that discards LBS-based deformation and synthesizes full-body clothing points from a global pose feature, as illustrated in \cref{fig:ablate_def}. The resulting noisy surfaces in articulated regions and overly coarse details further emphasize the necessity of the hybrid paradigm.

\begin{figure}[t]
  \centering
  \includegraphics[width=0.95\linewidth]{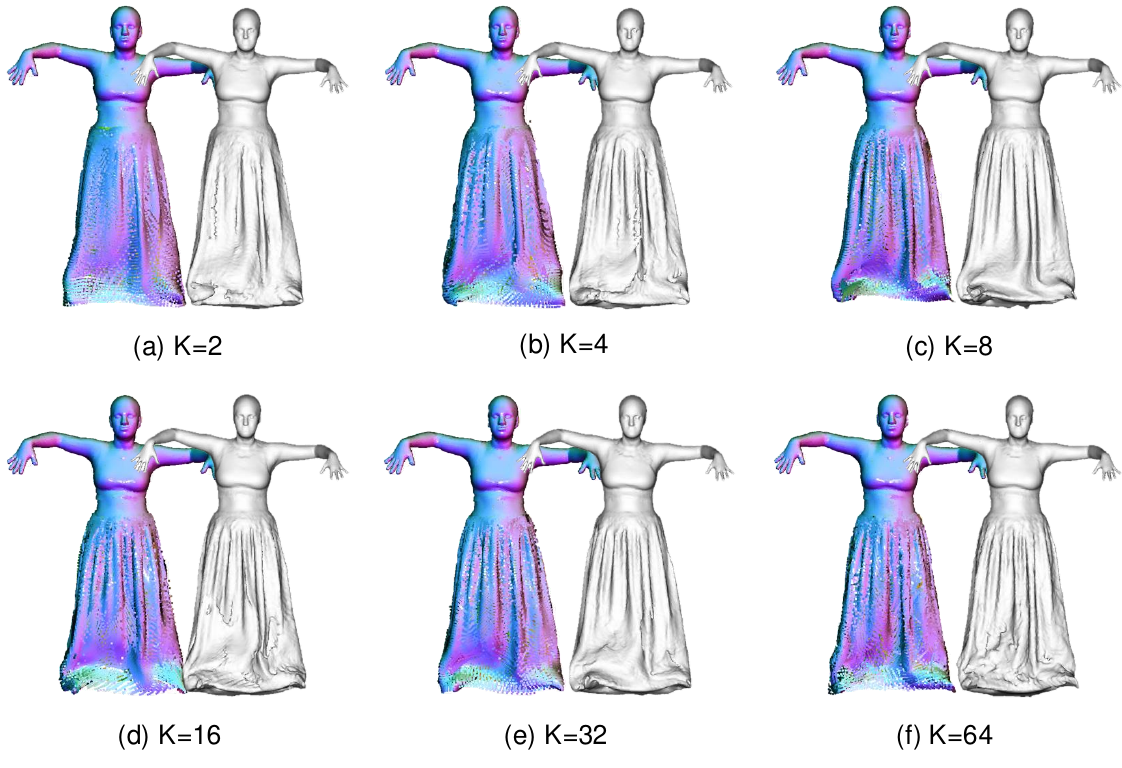}
  \vspace{-1ex}
  \caption{\textbf{Ablation studies of the number of patches $K$ used in the free-form generator.} The visualizations reveal that opting for $K=2$ leads to smoothed results without details. Empirically, selecting $K=8$ yields the best visual outcomes. In other cases, the surface becomes progressively noisier, compromising the clarity of fine-grained elements like clothing wrinkles.}
  \label{fig:supp_ablation_patch}
  \vspace{-1.em}
\end{figure}

\paragraph{Number of Patches.} 
We conduct ablation studies to manipulate the number of patches $K$ utilized in the free-form generator. This experiment also serves to illustrate the inherent complexity involved in modeling loose garments. As a case study, we select the long dress, which features intricate details such as wrinkles. The results, shown in \cref{fig:supp_ablation_patch}, are compared across different patch sizes: $K=2, 4, 8, 16, 32, 64$.

Visualization (a) shows that $K=2$ fails to capture the intricate details of the long dress, resulting in over-smoothed outcomes. In addition, empirical findings indicate that the setting $K=8$ is sufficient to generate high-quality details, producing superior visual results. When $K$ surpasses $8$ or is set to $4$, the generated surface exhibits increased noise, leading to a loss of clarity in fine-grained details. Notably, features such as clothing wrinkles become less distinct and sharply defined. This observation suggests that modeling the ostensibly complex long dress may be less daunting than anticipated. Furthermore, it verifies the remarkable expressiveness of our hybrid framework. Unless otherwise stated, $K=8$ is selected for our experiments. 

\begin{figure}[t]
  \centering
  \includegraphics[width=0.9\linewidth]{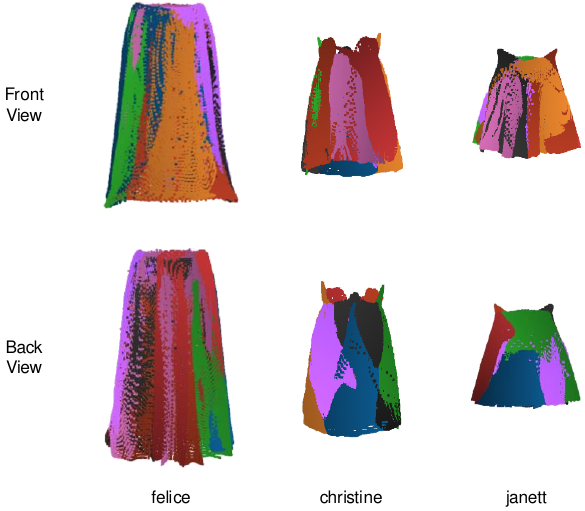}
  \vspace{-1.5ex}
  \caption{\textbf{Visualization of the generated $K=8$ patches which comprise the loose dress and skirts.} We show the results on the three subjects.}
  \label{fig:supp_vis_patch}
\end{figure}

To better investigate the properties of the free-form generator, we visualize $K=8$ patches of two loose skirts and a long dress, each rendered in different colors. As illustrated in \cref{fig:supp_vis_patch}, our free-form generator successfully recovers authentic fine-grained details while preserving good locality within each patch. Generally, the points within each patch are arranged in a vertical direction, and different patches seamlessly integrate to form a complete surface. These results, derived from data-driven learning, suggest that decomposing a loose garment into several ``vertical" patches is a plausible approach for detailed modeling. Additionally, we visualize the convergence process of the free-form generator throughout the training phase in the demo.

\begin{figure}[t]
  \centering
  \includegraphics[width=0.85\linewidth]{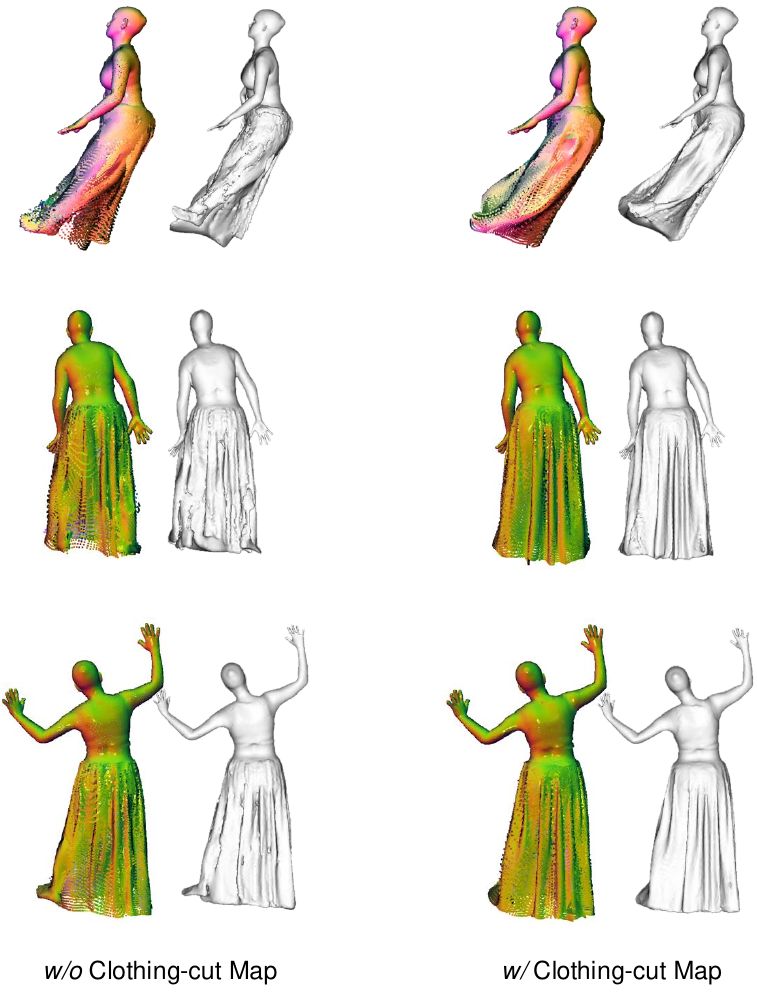}
  \vspace{-1.5ex}
  \caption{\textbf{More ablation study results of employing clothing-cut map.} Split-up artifacts between two modules can result in disjointed, noisy areas and torn appearances. The use of the clothing-cut map notably mitigates this issue.}
  \label{fig:ablation_supp_cut}
\end{figure}

\paragraph{Clothing-cut Map.} As discussed in the main paper, directly employing the generation module to fill in the loose regions in the previous LBS-based framework can cause split-up artifacts. Here we present additional cases demonstrating that this issue occurs under various poses, as illustrated in \cref{fig:ablation_supp_cut}. This issue becomes particularly evident when the underlying leg approaches the surface of the dress. In such instances, deformed points from the leg and the generated region become disjoint, causing the dress to appear torn. Additionally, a partial shape of the underlying leg can be observed through the cracks in the broken dress.

\subsection{Efficiency Analysis}
We evaluate the inference speed of our approach and other SOTA methods on an RTX 3090 GPU, with a batch size of 1. Additionally, we report the FLOPs and parameter counts to quantify computational resource requirements. As shown in \cref{tab:time}, our model has the smallest number of parameters and achieves real-time inference speed at $64.1$ FPS. Notably, our model significantly outperforms SOTAs without introducing extra computational overhead. 

\begin{table}[t]
    \caption{\textbf{Efficiency analysis of our method with other works.}}
    \label{tab:time}
    \vspace{-0.5em}
    \centering
    \resizebox{\linewidth}{!}{ %
    \setlength{\tabcolsep}{2mm}
    \begin{tabular}{ccccc}
        \hline
        Method & FID $\downarrow$ & FPS $\uparrow$ & FLOPS (G) & Params. (M) \\
        \hline
        POP~\cite{ma2021pop} & 57.87  & 69.9 & 128.81 & 11.33 \\
        SkiRT~\cite{ma2022skirt} & 53.32  & 79.9 & 77.12 & 11.13 \\
        FITE~\cite{lin2022fite} & 39.02 & 31.5 & 68.87 & 11.02 \\
        Ours & 37.75 & 64.1 & 78.82 & 10.83 \\
        \hline
    \end{tabular}}
\end{table}

\subsection{Limitations and Failure Cases}

Currently, our method focuses on single-frame modeling of clothed humans and does not consider the temporal cues that could provide constraints for clothing deformation due to motion. Consequently, discontinuities may appear in transitions between frames. Future work could explore incorporating temporal information to achieve smoother and more realistic modeling results.

\begin{figure}[h]
  \centering
  \includegraphics[width=0.9\linewidth]{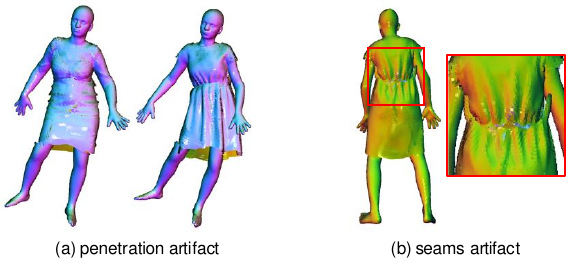}
   \vspace{-1ex}
  \caption{\textbf{Two typical failure modes.} (a) In challenging poses, the generated skirt or dress occasionally collides with the human body. (b) At the boundaries between deformed and generated regions, our model sometimes produces discontinuous "seams".} 
  \label{fig:failure_case}
\end{figure}

Despite employing pose augmentation, our model remains susceptible to failure when confronted with extremely challenging poses, resulting in clothing penetration artifacts, as depicted in \cref{fig:failure_case} (a). This issue is particularly noticeable when the skirt becomes tighter. Training our free-form generator on a larger dataset could enhance its robustness to out-of-distribution poses and reduce such artifacts. Additionally, while our pipeline employs different strategies to handle deformed and generated areas, we cannot guarantee the perfect blending of point clouds from two modules. As illustrated in \cref{fig:failure_case} (b),  ``seams'' at the boundary regions are occasionally observed.

However, it is crucial to note that our experiments underscore the promise and versatility of our proposed hybrid approach. By transcending the limitations imposed by relying solely on LBS-based deformation, our method demonstrates notable expressive capabilities. We believe that with larger datasets, our approach has considerable potential for superior performance in future applications.

\clearpage

\clearpage

\clearpage

\clearpage

\clearpage

\clearpage

\end{document}